\documentclass{article} 
\usepackage{robotpolicy_preprint,times}

\usepackage[T1]{fontenc}
\usepackage{amsmath,amssymb}
\usepackage{graphicx,xcolor,booktabs,tabularx,array,longtable,multirow}
\usepackage{pgfplots}
\usetikzlibrary{arrows.meta,positioning,calc}
\usepackage{placeins}
\usepackage{float}
\pgfplotsset{compat=1.18}
\usepackage{hyperref}
\usepackage{url}
\hypersetup{hidelinks,
  pdftitle={What Should Robot Policies Learn to Predict? Rethinking Supervision Targets in Vision-Language-Action Learning}}

\newif\ifincludeappendix
\includeappendixtrue

\newcolumntype{Y}{>{\centering\arraybackslash}X}

\title{\hyphenpenalty=10000\exhyphenpenalty=10000 What Should Robot Policies Learn to Predict? Rethinking Supervision Targets in Vision-Language-Action Learning}
\author{%
  {\bfseries
  Haitao Shen\textsuperscript{1}\thanks{Equal contribution.}\quad Hanyu Liu\textsuperscript{2}\footnotemark[1]\quad
  Yihan Lin\textsuperscript{1}\quad Huiping Liu\textsuperscript{3}\quad
  Junbo Cui\textsuperscript{4}\\
  Daoheng Li\textsuperscript{5}\quad Yang Li\textsuperscript{1}\quad
  Jing Zhang\textsuperscript{1}\quad Wenxin Li\textsuperscript{2}\quad
  Yuan Yao\textsuperscript{4,5}}\\
  \textsuperscript{1}Renmin University of China\quad
  \textsuperscript{2}Peking University\\
  \textsuperscript{3}Beijing Jiaotong University\quad
  \textsuperscript{4}ModelBest\\
  \textsuperscript{5}Tsinghua University%
}
\hypersetup{pdfauthor={Haitao Shen, Hanyu Liu, Yihan Lin, Huiping Liu, Junbo Cui, Daoheng Li, Yang Li, Jing Zhang, Wenxin Li, Yuan Yao}}

\begin{document}
\maketitle
\begin{abstract}
Vision-language-action (VLA) policies are typically trained with action
prediction as the primary supervision signal, since robots ultimately execute
actions. Auxiliary objectives such as latent-action (LA) and future-vision (FV)
prediction are often added to improve policy learning, yet they compete with
action prediction for the same training budget. We ask whether, under a fixed
budget, allocating a substantial share of training samples to such objectives,
rather than to further action prediction, can improve policy learning. We build
a unified VLA framework that couples a two-stream multimodal diffusion
transformer (MMDiT) with a sample-level router, which assigns each training
sample to either action prediction or an auxiliary objective and thus makes
supervision allocation explicit. Within this framework, we instantiate three LA
targets based on LAPA and its DINO-feature variants, and three FV targets based
on DINOv3, V-JEPA~2, and Wan2.2 representations. Across LIBERO, CALVIN, and
four real-robot tasks, suitable LA and FV supervision outperforms action-only
training under the same budget, even when action prediction receives no more
than half of the training samples. FV generally outperforms LA, although the
strongest LA target remains competitive on CALVIN. Diagnostic analyses suggest
that DINO-based LA targets emphasize relative motion, whereas FV targets
emphasize absolute position and task identity. We therefore argue that what
robot policies learn to predict, and how training is allocated among these
targets, should be treated as explicit design choices rather than defaulting
to action-dominant training.
\end{abstract}

\section{Introduction}
\label{sec:intro}

Vision-language-action (VLA) models have emerged as a prominent paradigm for robot learning, achieving strong performance across many manipulation tasks~\citep{baseline_openvla,baseline_pi05}. Beyond direct action supervision, recent studies increasingly introduce additional learning objectives into VLA training, including geometric representation alignment and spatial or 3D structure prediction~\citep{intro_spatial_forcing,intro_glad}. Among these, latent-action (LA) and future-vision (FV) prediction have become two representative forms of supervision. Latent action abstracts visual transitions into compact tokens that characterize how the scene changes over time~\citep{intro_lapa,intro_villax}, whereas future vision encourages the policy to predict future representations, providing supervision about how the scene is expected to evolve~\citep{intro_gr1,rw_flare}.

\begin{figure*}[!t]
\centering
\includegraphics[width=\textwidth]{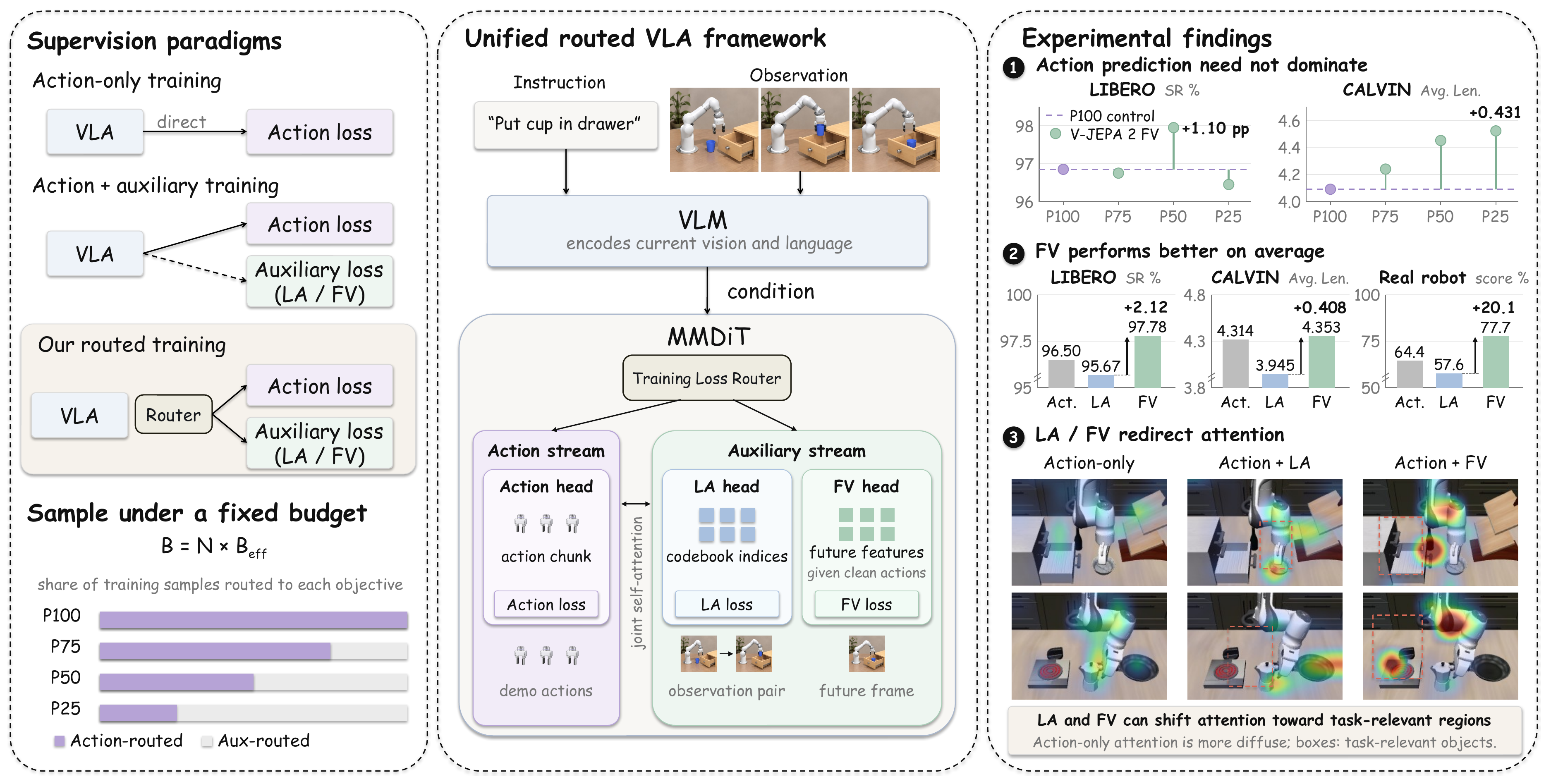}
\caption{\textbf{Overview.} \textbf{Left:} under a fixed training budget, a router
sends each sample to the action loss or one auxiliary loss (LA or FV); P$x$ routes
$x$\% of samples to the action loss. \textbf{Middle:} two-stream VLA; FV prediction
conditions on clean demonstrated actions. \textbf{Right:} (1) V-JEPA~2 FV vs.\ its
source-matched P100 control; (2) LA and FV means over three targets each at P50
vs.\ action-only (Act.), arrows marking unrounded FV$-$LA (real robot: \% of max
score); (3) per-map normalized LIBERO attention for action-only, +LA (LAPA-DINOv3),
and +FV (V-JEPA~2).}
\label{fig:framework}
\end{figure*}

Because executable actions are the ultimate outputs of robot policies, direct action prediction is typically treated as the dominant training objective, while LA and FV prediction are introduced as auxiliary supervision. Prior studies have examined latent-action formulations and integration strategies~\citep{intro_pixels_to_tokens,intro_what_matters}, showing that such supervision can benefit downstream policy learning. More importantly, LA and FV are not merely auxiliary in effect: by providing learning signals about state changes and task structure, they directly shape the same policy during optimization. Yet, showing that an additional target is useful does not establish that direct action prediction is itself the most effective supervision. We therefore ask (Figure~\ref{fig:framework}): \emph{under a fixed training budget, what should robot policies learn to predict, and can assigning LA or FV prediction an equal or larger share of training samples than direct action prediction improve action execution?}

To investigate this question, we build a unified VLA framework that places a two-stream MMDiT on top of a pretrained VLM, with separate action and auxiliary experts interacting through joint self-attention. We further introduce a sample-level router that assigns each training sample to either action or auxiliary supervision under a fixed training budget, making their relative training allocation explicit. Within this framework, we instantiate three LA targets and three FV targets: LA predicts discrete latent tokens extracted by different encoders, whereas FV predicts future features in different visual representation spaces. This unified setup enables a controlled comparison of how different supervision targets shape policy learning, with source-matched controls that change only routing for each FV target; FV prediction still conditions on demonstrated actions.

We evaluate our framework on two simulation benchmarks (LIBERO and CALVIN) and four real-robot manipulation tasks, and obtain three main findings. First, \textbf{suitable LA and FV supervision can outperform action-only training} under the same training budget, even when direct action prediction receives only half or fewer of the training samples. Second, \textbf{FV generally performs better than LA} across both simulation and real-robot tasks, while the strongest LA configuration remains competitive on CALVIN. Third, our analyses suggest that \textbf{LA and FV shape policy learning in different ways}: DINO-based LA targets emphasize relative motion, whereas FV targets better capture absolute position and task identity, and both can alter policy attention toward task-relevant regions.

Our contributions are threefold: \textbf{(1)} We revisit the dominant role of direct action prediction in VLA training and investigate how allocating supervision to LA and FV affects policy learning. \textbf{(2)} We introduce a unified VLA training framework with a sample-level training router that explicitly controls the allocation of action and auxiliary supervision under a fixed training budget, enabling controlled comparisons across multiple LA and FV targets. \textbf{(3)} We conduct extensive experiments on LIBERO, CALVIN, and four real-robot manipulation tasks, together with representation and attention analyses, and derive systematic findings on action-supervision allocation, the performance differences between LA and FV, and their representation characteristics.

\section{Related Work}
\label{sec:related}

\paragraph{Supervision beyond action labels.}
Robot learning draws on supervision beyond action labels, including visual
reconstruction and language generation~\citep{rw_voltron}, vision-language
co-training~\citep{rw_rt2}, and geometric
alignment~\citep{intro_spatial_forcing,intro_glad,rw_4d_wam}. We focus on two
demonstration-derived targets, LA and FV, and study both which target to
predict and how to allocate training samples between it and direct action
prediction.

\paragraph{Latent actions.}
LAPA and UniVLA pretrain policies with latent actions derived from visual
transitions~\citep{intro_lapa,rw_univla}. The villa-X framework jointly models
latent and robot actions, using latent plans to support action
generation~\citep{intro_villax}. Analyses examine controllable changes and
exogenous noise~\citep{rw_lam_learn}, while LARYBench~\citep{rw_lary} finds that
general visual encoders can outperform specialized latent-action models on its
semantic and physical-action probes. Empirical studies compare latent-action
formulations, integration strategies, and loss weights~\citep{intro_pixels_to_tokens},
and examine links between representation-quality proxies and downstream policy
performance~\citep{intro_what_matters}. Rather than reweighting losses on every
sample, we route each sample to a single objective, and we probe LA and FV
sources with the same absolute- and relative-motion readouts.

\paragraph{Future vision.}
GR-1, GR-2, UWM, and DreamZero jointly model future images or videos and robot
actions~\citep{intro_gr1,rw_gr2,rw_uwm,rw_dreamzero}. Seer and VPP guide action
generation with predicted visual states or features~\citep{rw_seer,intro_vpp}.
FLARE enhances policy learning through future-representation alignment and
compares against an action-only baseline trained for longer~\citep{rw_flare}.
We instead compare against action-only training with the same total budget,
vary the action share, and add source-matched FV controls, including
allocations where auxiliary prediction receives most samples.

\section{Method}
\label{sec:supervision_selection}

\subsection{Problem Setup and Supervision Targets}
\label{sec:predictive_targets}

\paragraph{Problem setup.}
 At time \(t\), the policy receives the current visual observation \(o_t\), the language instruction \(\ell\), and the robot state \(s_t\), and predicts an action chunk over the next \(H\) steps:
\begin{equation}
\mathbf a_t=(a_t,\ldots,a_{t+H-1}),
\qquad
\hat{\mathbf a}_t\sim\pi_\theta(\cdot\mid o_t,\ell,s_t).
\end{equation}
We study the allocation of supervision under a fixed training budget. Suppose training runs for \(N\) optimizer updates with effective batch size \(B_{\mathrm{eff}}\). For a selected auxiliary prediction target \(q\in\{\mathrm{LA},\mathrm{FV}\}\), we define the total training budget and its allocation between action and auxiliary prediction as \(\mathcal B=NB_{\mathrm{eff}}=\mathcal B_A+\mathcal B_q\).
Here \(\mathcal B\) counts training-sample occurrences, including repeated uses of the same demonstration. Matching this budget fixes sample exposure; it does not establish equal FLOPs or wall-clock time.

\paragraph{Latent action (LA).}
 We encode each visual transition into a group of discrete LA tokens using an LA encoder \(E_{\mathrm{LA}}^{(m)}\):
\begin{equation}
c_u=E_{\mathrm{LA}}^{(m)}(o_u,o_{u+\delta}),
\qquad m\in\{\mathrm{LAPA},\mathrm{LAPA\text{-}DINOv2},
\mathrm{LAPA\text{-}DINOv3}\}.
\end{equation}
Here \(\delta\) is the temporal offset between the observations used to construct a transition label, and \(c_u=(c_{u,1},\ldots,c_{u,K})\) contains \(K\) discrete codebook indices. The policy predicts the concatenation \(z_t=(c_t,\ldots,c_{t+H-1})\), flattened into \(L_z=HK\) token positions, so that the target spans the same \(H\) steps as the action chunk. We use \(H=8\); the benchmark-specific temporal construction is given in Appendix~\ref{app:available_configuration}. LAPA encodes raw-image changes, whereas LAPA-DINOv2 and LAPA-DINOv3 (collectively, LAPA-DINO) encode changes between DINOv2 and DINOv3 visual representations. The same appendix gives token counts and codebooks.

\paragraph{Future vision (FV).}
 We use an FV encoder \(E_{\mathrm{FV}}^{(m)}\) to encode the current and future visual observations:
\begin{equation}
f_t=E_{\mathrm{FV}}^{(m)}(o_t),
\qquad
f_{t+\delta}=E_{\mathrm{FV}}^{(m)}(o_{t+\delta}),
\qquad m\in\{\mathrm{DINOv3},\mathrm{V\text{-}JEPA\ 2},
\mathrm{Wan2.2\ VAE}\}.
\end{equation}
Both \(f_t\) and \(f_{t+\delta}\) are continuous visual representations. We use \(f_t\) as an MMDiT input and construct the FV flow-matching target from \(f_{t+\delta}\), with \(\delta=8\) for FV. Detailed input formats and feature dimensions for each encoder are provided in Appendix~\ref{app:available_configuration}. For both LA and FV, we refer to the encoder \(m\) as the \emph{source} of the target.

\subsection{Policy Architecture and Routing}
\label{sec:two_stream_policy}

\begin{figure*}[!t]
\centering
\includegraphics[width=0.88\textwidth]{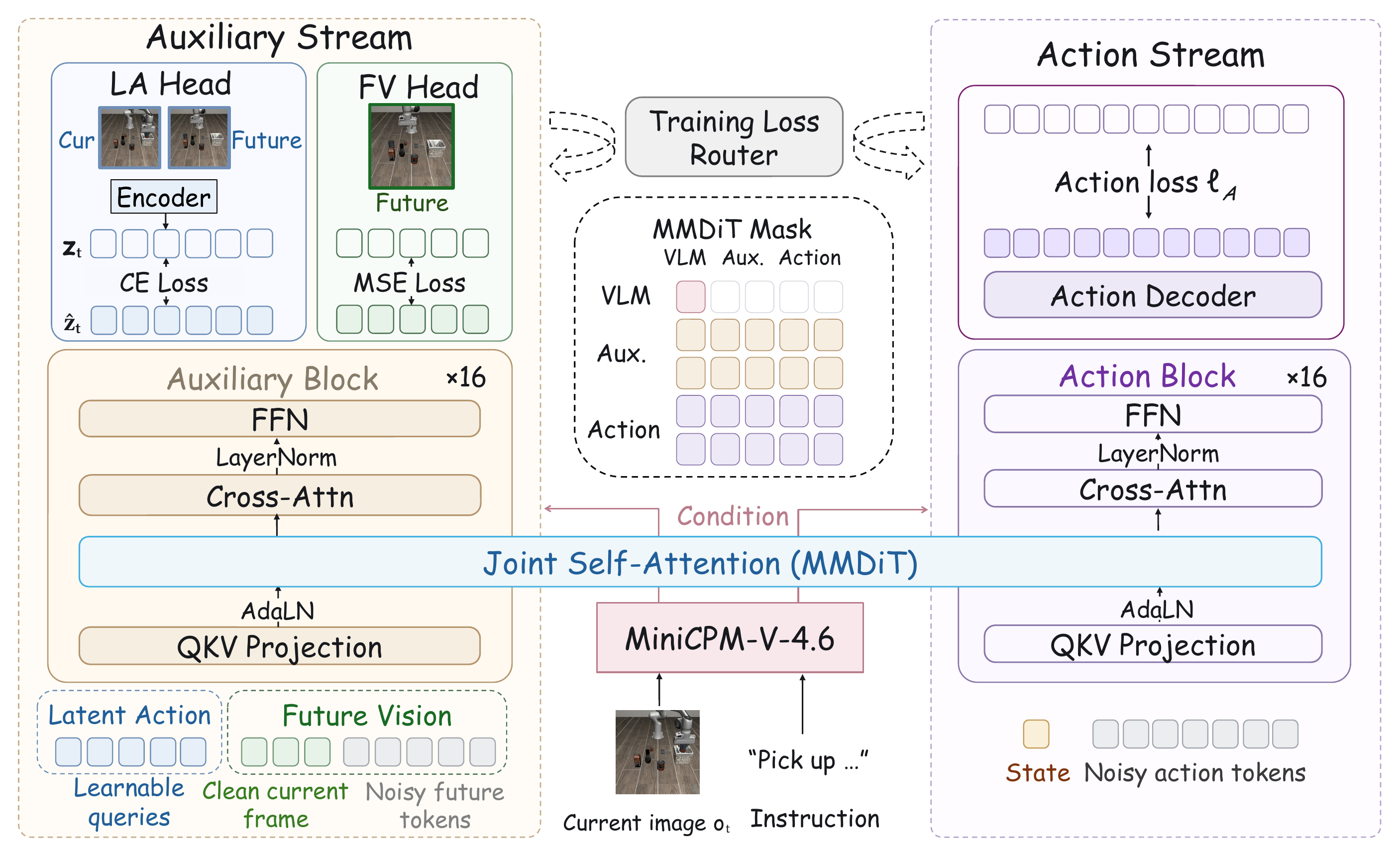}
\caption{VLM--MMDiT architecture. Action and auxiliary (LA/FV) experts interact through
joint self-attention; sample-level routing selects the active loss.
Future observations are used only for auxiliary training.}
\label{fig:architecture}
\end{figure*}

\paragraph{VLM conditioning.}
 We use a pretrained VLM to encode the current visual observation and language instruction as \(h_t=\operatorname{VLM}_{\phi}(o_t,\ell)\),
where \(h_t\) denotes the last hidden states of the VLM. We use MiniCPM-V-4.6 and condition both MMDiT experts on \(h_t\) through cross-attention, while \(s_t\) separately conditions the action expert.

\paragraph{MMDiT prediction head.}
 The VLM is followed by a two-stream MMDiT prediction head (Figure~\ref{fig:architecture}) consisting of an action expert and an auxiliary expert. The two experts use separate input projections and feed-forward networks and exchange information through joint self-attention.

\paragraph{Sample-level training router.}
 To allocate supervision across predictive targets under a fixed training budget, we route each training sample to either the action prediction objective or the selected LA or FV prediction objective. The router determines the active loss and the MMDiT input construction, which differ between routes (e.g., FV-routed samples condition on clean demonstrated actions, whereas action-routed samples receive noisy actions); each sample therefore contributes a single loss rather than a weighted sum of both, and each sample occurrence counts toward exactly one of \(\mathcal B_A\) and \(\mathcal B_q\). Both experts participate in joint self-attention on either route.
For a selected auxiliary target \(q\in\{\mathrm{LA},\mathrm{FV}\}\), we independently sample a routing indicator \(r_i\) for each training example. The routing distribution and expected allocations under the fixed budget \(\mathcal B\) are
\begin{equation}
r_i\sim\operatorname{Bernoulli}(p_A),
\qquad
\mathbb E[\mathcal B_A]=p_A\mathcal B,
\qquad
\mathbb E[\mathcal B_q]=(1-p_A)\mathcal B.
\end{equation}
Here \(r_i=1\) denotes the action route and \(r_i=0\) denotes the auxiliary route. We denote \(p_A=0.25,\ 0.50,\ 0.75,\) and \(1.00\) as P25, P50, P75, and P100, respectively. These proportions count only samples assigned to the direct action loss, not all uses of action labels (FV-routed samples also condition on demonstrated actions) or relative gradient magnitudes. For each FV configuration, the \emph{source-matched control} is its P100 version: it keeps the same FV encoder, MMDiT input \(f_t\), two-stream architecture, and training settings, but every sample follows the action route, so the FV loss is never active.

\paragraph{Auxiliary expert configurations.}
 The auxiliary expert is instantiated differently for LA and FV supervision. For LA, we introduce \(L_z\) learnable query tokens \(Q_{\mathrm{LA}}\), each corresponding to one LA token position. The LA expert predicts
\begin{equation}
\hat z_t\sim\pi_\theta^{\mathrm{LA}}
(\cdot\mid Q_{\mathrm{LA}},h_t),
\end{equation}
where \(z_t\) is the concatenated LA target defined in Section~\ref{sec:predictive_targets}. Since \(Q_{\mathrm{LA}}\) carries no source-specific features, LA configurations use the shared full-budget action-only baseline as their P100 reference.
For FV, the auxiliary expert takes the current visual representation \(f_t\) together with a noisy future visual representation \(f_{t+\delta}^{\tau_f}\) and the action \(\mathbf a_t\), and predicts the flow velocity of the future visual representation:
\begin{equation}
\hat v_{t+\delta}^{f}
=
v_\theta^{\mathrm{FV}}
\left(
f_t,f_{t+\delta}^{\tau_f},\mathbf a_t,h_t,\tau_f
\right),
\end{equation}
where \(\tau_f\in[0,1]\) denotes the flow time. On FV-routed training samples, \(\mathbf a_t\) is the clean demonstrated action chunk. The FV expert is trained with flow matching (Section~\ref{sec:training_objectives}).

\paragraph{Action expert.}
 The action expert takes the robot state \(s_t\) and a noisy action chunk \(\mathbf a_t^{\tau_a}\), and predicts the clean action chunk conditioned on the VLM hidden states \(h_t\):
\begin{equation}
\hat{\mathbf a}_t\sim\pi_\theta^{A}
\left(
\cdot\mid\mathbf a_t^{\tau_a},s_t,h_t,\tau_a
\right),
\end{equation}
where \(\tau_a\in[0,1]\) is the action noise time and \(\mathbf a_t^{\tau_a}\) is the corresponding noisy action input.
At inference, we use the same action-route computation as for action-routed training samples. Both MMDiT experts participate in joint self-attention, but no future observation or auxiliary target is provided, and only the action-expert output is decoded and executed.

\subsection{Training Objectives}
\label{sec:training_objectives}

\paragraph{Noise construction.}
 For action and FV prediction, we sample independent Gaussian noise \(\epsilon_a,\epsilon_f\sim\mathcal N(0,I)\). With \(\tau_a,\tau_f\in[0,1]\), the noisy inputs and FV flow velocity are
\begin{equation}
\begin{aligned}
\mathbf a_t^{\tau_a}
&=(1-\tau_a)\epsilon_a+\tau_a\mathbf a_t,\\
f_{t+\delta}^{\tau_f}
&=(1-\tau_f)\epsilon_f+\tau_f f_{t+\delta},
\qquad
v_{t+\delta}^{f}=f_{t+\delta}-\epsilon_f.
\end{aligned}
\end{equation}

\paragraph{Prediction losses.}
 The action objective and auxiliary objectives are
\begin{equation}
\mathcal L_A=\ell_A(\hat{\mathbf a}_t,\mathbf a_t),
\qquad
\mathcal L_{\mathrm{LA}}=\operatorname{CE}(\hat z_t,z_t),
\qquad
\mathcal L_{\mathrm{FV}}
=
\left\|\hat v_{t+\delta}^{f}-v_{t+\delta}^{f}\right\|_2^2.
\label{eq:policy_objectives}
\end{equation}
Here \(\ell_A\) is the action loss defined by each training dataset, whose clean-action parameterization and reduction are specified in Appendix~\ref{app:protocol_records}; \(\mathcal L_{\mathrm{LA}}\) is averaged over valid LA tokens.

\paragraph{Routed objective.}
 Each sample contributes only the loss associated with its selected route. For a batch of \(B\) samples, the overall training objective is
\begin{equation}
\mathcal L
=
\frac{1}{B}
\sum_{i=1}^{B}
\left[
r_i\mathcal L_{A,i}
+
(1-r_i)\lambda_q\mathcal L_{q,i}
\right],
\qquad
q\in\{\mathrm{LA},\mathrm{FV}\},
\end{equation}
where \(\lambda_q\) is a target-specific loss weight, tuned for LA and fixed across allocations (Appendix~\ref{app:la_weight_ablation}).

\section{Experiments}
\label{sec:policy}
We investigate the roles of LA and FV in policy learning through
simulation benchmarks, real-robot experiments, and representation
analyses, focusing on three research questions:

\noindent\emph{Question~1. Must direct action prediction receive most supervision?}

\noindent\emph{Question~2. How do LA and FV compare across training regimes and tasks?}

\noindent\emph{Question~3. Why can LA and FV supervision help policy learning?}

\subsection{Experimental Setup}
\label{sec:exp_setup}

\paragraph{Benchmarks.}
\label{sec:benchmarks_evaluation}
We evaluate on LIBERO~\citep{benchmark_libero} across its Spatial, Object,
Goal, and Long suites using task success rates. On
CALVIN~\citep{benchmark_calvin}, we follow ABC$\rightarrow$D and report
average task-chain length (Avg. Len.) and five-task success (SR-5).
Evaluation protocols are detailed in Appendix~\ref{app:calvin_metrics}.
We also evaluate four real-robot tasks involving basket pick-and-place
and cup/object rearrangement, scored by completion progress.

\paragraph{Training configurations.}
\label{sec:training_controls}
All policies use MiniCPM-V-4.6 in the unified framework
(Section~\ref{sec:supervision_selection}). On LIBERO and CALVIN, the
\emph{full budget} is 60k optimizer updates and the \emph{half budget} is
30k; results use the full budget unless noted. The full-budget action-only
baseline spends the entire budget on action prediction, and the
half-budget baseline matches the expected number of action-routed samples
in a full-budget P50 run. Detailed configurations and evaluation protocols are provided in
Appendices~\ref{app:implementation}
and~\ref{app:beyond_standard_evaluations}.

\subsection{Must Direct Action Prediction Receive Most Supervision?}
\label{sec:ablations}
\begingroup

We first evaluate an equal allocation between action and auxiliary
prediction (P50; Table~\ref{tab:policy}). Against the half-budget
action-only baseline, which receives the same expected number of action
samples, all six configurations improve LIBERO success, and all FV
configurations and LAPA-DINOv3 improve CALVIN average task-chain length,
whereas LAPA and LAPA-DINOv2 do not. Against the full-budget baseline,
which spends the entire budget on action prediction, all three FV
configurations exceed its 96.50\% on LIBERO (up to 98.15\% with Wan2.2), and
DINOv3, V-JEPA~2, and LAPA-DINOv3 exceed its 4.314 on CALVIN (up to 4.451
with V-JEPA~2). Suitable configurations can thus
outperform action-only training while assigning only half of their samples
to direct action prediction, although no LA configuration does so on
LIBERO. Since configurations also differ in visual inputs and auxiliary experts, we next examine allocation with the corresponding controls.

\begin{table}[!t]
\centering
\nocite{baseline_openvla,baseline_pi0,rw_univla,baseline_pi05}
\caption{LIBERO and CALVIN policy performance. Mixed configurations use P50
and the full budget. Parentheses: differences from the half-budget baseline. S/O/G/L: Spatial/Object/Goal/Long;
``--'': unavailable. Source-matched FV controls: Appendices~\ref{app:policy_checkpoints}, \ref{app:calvin_checkpoints}, and~\ref{app:calvin_allocations}.}
\label{tab:policy}
\begingroup
\small
\setlength{\tabcolsep}{0.8pt}
\renewcommand{\arraystretch}{1.0}
\newcommand{\posdelta}[1]{{\textcolor{green!45!black}{(#1)}}}
\newcommand{\negdelta}[1]{{\textcolor{red!65!black}{(#1)}}}
\begin{tabular*}{\linewidth}{@{\extracolsep{\fill}}@{}llcrrrrr@{\hspace{6pt}}rr@{}}
\toprule
& \multirow{2}{*}{Method} & \multirow{2}{*}{\shortstack{Param.\\(B)}}
& \multicolumn{5}{c}{LIBERO}
& \multicolumn{2}{c}{CALVIN} \\
\cmidrule(lr){4-8}\cmidrule(l){9-10}
& & & S & O & G & L & Avg. $\uparrow$
& SR-5 $\uparrow$ & Avg. Len. $\uparrow$ \\
\midrule
\multirow{4}{*}{\textit{Ref.}} & OpenVLA & $\approx\!8$ & 84.7 & 88.4 & 79.2 & 53.7 & 76.5 & 43.5 & 3.270 \\
& $\pi_0$ & $\approx\!3$ & 96.8 & 98.8 & 95.8 & 85.2 & 94.2 & 60.1 & 3.840 \\
& UniVLA & $\approx\!7$ & 96.5 & 96.8 & 95.6 & 92.0 & 95.2 & 56.5 & 3.800 \\
& $\pi_{0.5}$ & $\approx\!3$ & 98.8 & 98.2 & 98.0 & 92.4 & 96.9 & -- & -- \\
\midrule
\multirow{2}{*}{\textit{Act.}} & Half budget & $\approx\!2$ & 97.2 & 99.4 & 94.6 & 81.6 & 93.20 & 57.7 & 3.971 \\
& Full budget & $\approx\!2$ & 98.2 & 98.6 & 98.2 & 91.0
& 96.50\,\posdelta{$+3.30$}
& 70.3\,\posdelta{$+12.6$}
& 4.314\,\posdelta{$+0.343$} \\
\midrule
\multirow{3}{*}{\textit{LA}} & LAPA & $\approx\!2$ & 96.4 & 98.2 & 94.2 & 89.8
& 94.65\,\posdelta{$+1.45$}
& 42.1\,\negdelta{$-15.6$}
& 3.466\,\negdelta{$-0.505$} \\
& LAPA-DINOv2 & $\approx\!2$ & 97.0 & 99.6 & 98.0 & 91.2
& 96.45\,\posdelta{$+3.25$}
& 63.4\,\posdelta{$+5.7$}
& 3.935\,\negdelta{$-0.036$} \\
& LAPA-DINOv3 & $\approx\!2$ & 97.2 & 97.2 & 96.8 & 92.4
& 95.90\,\posdelta{$+2.70$}
& \textbf{77.0}\,\posdelta{$+19.3$}
& 4.434\,\posdelta{$+0.463$} \\
\midrule
\multirow{3}{*}{\textit{FV}} & DINOv3 & $\approx\!2$ & 98.4 & 99.2 & 98.4 & 93.0
& 97.25\,\posdelta{$+4.05$}
& 75.1\,\posdelta{$+17.4$}
& 4.370\,\posdelta{$+0.399$} \\
& V-JEPA~2 & $\approx\!2$ & 98.8 & 99.2 & 98.6 & 95.2
& 97.95\,\posdelta{$+4.75$}
& 76.0\,\posdelta{$+18.3$}
& \textbf{4.451}\,\posdelta{$+0.480$} \\
& Wan2.2 & $\approx\!2$ & 98.4 & 100.0 & 99.0 & 95.2
& \textbf{98.15}\,\posdelta{$+4.95$}
& 70.5\,\posdelta{$+12.8$}
& 4.237\,\posdelta{$+0.266$} \\
\bottomrule
\end{tabular*}
\endgroup
\end{table}

Figure~\ref{fig:target_allocation} varies the action share at a fixed
total budget. Each FV configuration uses a source-matched P100 control
that differs only in routing; LA uses the shared full-budget action-only
baseline. On CALVIN, V-JEPA~2 at P25 reaches an average task-chain length
of 4.522 versus 4.091 for its source-matched control, also exceeding the
full-budget baseline in Table~\ref{tab:policy}. This configuration
assigns 75\% of training samples to FV prediction in
expectation, showing that direct action prediction need not receive most
samples to improve task performance.

\begin{figure}[!htbp]
\centering
\includegraphics[width=\textwidth]{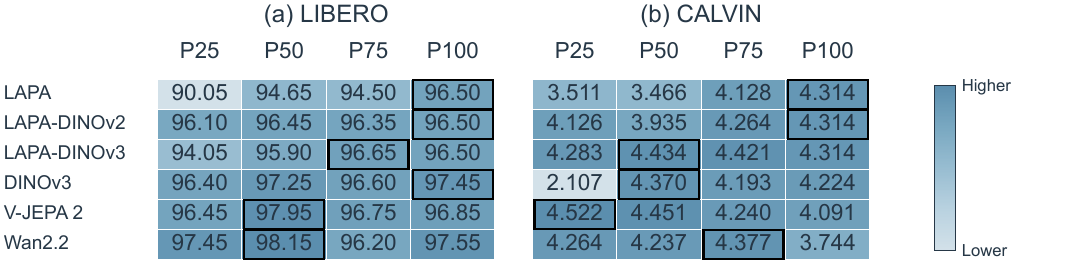}
\caption{Allocation at the full budget. Cells show LIBERO success (\%) and
CALVIN Avg. Len.; darker is higher within each benchmark. Black outlines
mark each model's best allocation.}
\label{fig:target_allocation}
\end{figure}

This does not imply that reducing action prediction always helps. On
LIBERO, V-JEPA~2 and Wan2.2 outperform their source-matched controls at P50,
but every tested P25 configuration falls below its P100 control. The
appropriate allocation therefore depends on both the target and the
benchmark.

Figure~\ref{fig:checkpoints} fixes P50 to
examine whether the gains persist across training budgets. On CALVIN,
all three FV configurations exceed their source-matched controls in
average task-chain length at all four tested budgets. These controls,
however, score below the shared action-only baseline at every budget, so
gains over that baseline are less consistent: V-JEPA~2 exceeds it at
three of the four budgets, and among LA configurations only LAPA-DINOv3
does so, at the full budget. On LIBERO, gains vary with the target and
budget and do not occur at every stage. Detailed LIBERO and CALVIN scores
are provided in Appendices~\ref{app:libero_route05_records}
and~\ref{app:calvin_checkpoints}.

\begin{figure}[!htbp]
\centering
\makebox[\textwidth][l]{\small\textbf{(a) CALVIN}}\par
\includegraphics[width=\textwidth]{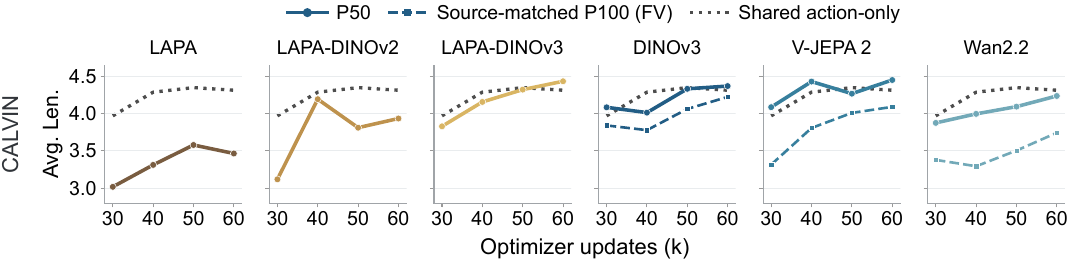}\par
\vspace{5pt}
\makebox[\textwidth][l]{\small\textbf{(b) LIBERO}}\par
\includegraphics[width=\textwidth,trim=0 0 0 14bp,clip]{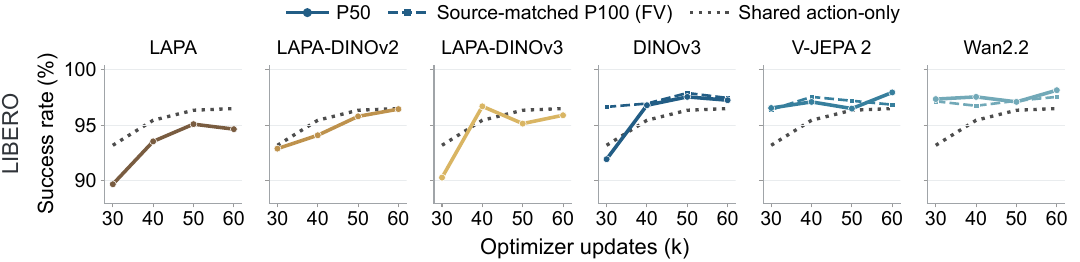}
\caption{Performance across training budgets at P50 on (a) CALVIN and
(b) LIBERO. Both panels share the line styles shown above: solid, P50;
colored dashed, source-matched FV P100; gray dotted, shared action-only
baseline, also the LA P100 reference.}
\label{fig:checkpoints}
\end{figure}

Figure~\ref{fig:allocation_budget_full} jointly varies supervision
allocation and training budget, reporting all three mixed allocations at
each tested budget on both benchmarks. LA uses the shared action-only
baseline at the same checkpoint; FV uses source-matched P100 controls.
This full sweep complements the fixed-full-budget allocation comparison
in Figure~\ref{fig:target_allocation} and the fixed-P50 budget comparisons
in Figure~\ref{fig:checkpoints}.
The plotted gains describe improvements within each configuration;
absolute scores and both reference types in the preceding figures
support comparisons across configurations. CALVIN gains over
source-matched controls for V-JEPA~2 and Wan2.2 extend across all tested
mixed allocations and budgets.

\begin{figure}[!htbp]
\centering
\includegraphics[width=\textwidth,trim=0 0 0 4bp,clip]{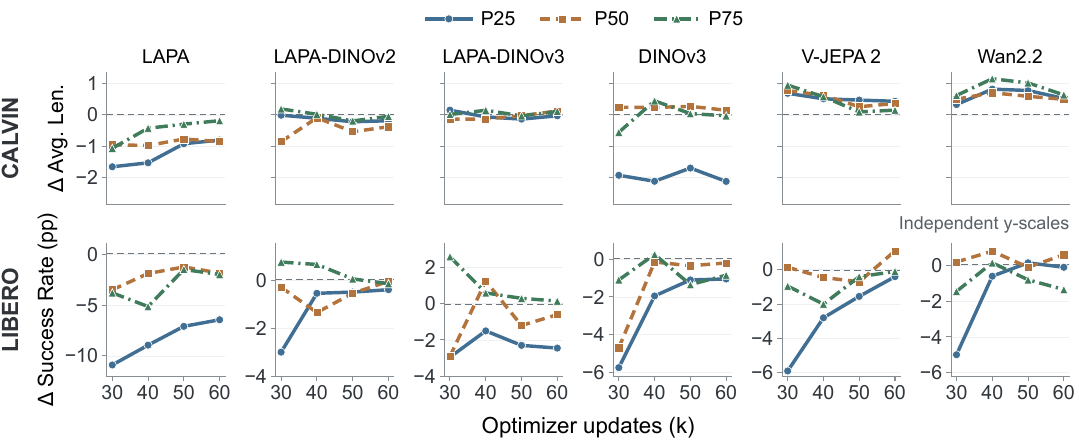}
\caption{Full sweep over supervision allocation and training budget.
Each curve reports gains over the same-stage P100 control: the shared
action-only baseline for LA and the source-matched control for FV.
Rows show CALVIN $\Delta$ Avg. Len. and LIBERO $\Delta$ success rate (pp).
Dashed zero lines indicate parity with the corresponding control.
CALVIN panels share a vertical scale; LIBERO panels use target-specific
vertical scales to show each trajectory.}
\label{fig:allocation_budget_full}
\end{figure}

\FloatBarrier
\endgroup

\subsection{How Do LA and FV Compare Across Training Regimes and Tasks?}
\label{sec:supervision_characteristics}

We next compare the complete LA and FV configurations, which differ
in target sources and auxiliary experts but share the action objective.

On LIBERO, FV leads overall. At P50 and the full budget
(Table~\ref{tab:policy}), all three FV configurations outperform all three
LA configurations in average success and on Spatial, Goal, and Long, while
scores overlap on Object. The highest LIBERO score at each tested mixed
allocation also comes from FV (Figure~\ref{fig:target_allocation}), and FV
generally scores higher than LA across the four P50 budgets
(Figure~\ref{fig:checkpoints}(b)). On CALVIN, the comparison depends on
allocation and budget. At P50 with the full budget, LAPA-DINOv3 nearly matches
the best FV configuration, V-JEPA~2 (4.434 vs.\ 4.451), and their order
reverses on SR-5 (77.0\% vs.\ 76.0\%). In average task-chain length, V-JEPA~2 leads at P25 and P50,
whereas LAPA-DINOv3 leads at P75 (4.421). Across P50 budgets
(Figure~\ref{fig:checkpoints}(a)), all FV configurations have longer average
task-chain lengths than all LA configurations at the half budget, but LAPA-DINOv3
becomes competitive with the best FV configurations at 50k updates and the full budget.

\begin{figure}[!htbp]
\centering
\includegraphics[width=\textwidth,trim=0 124.2bp 0 2.5bp,clip]{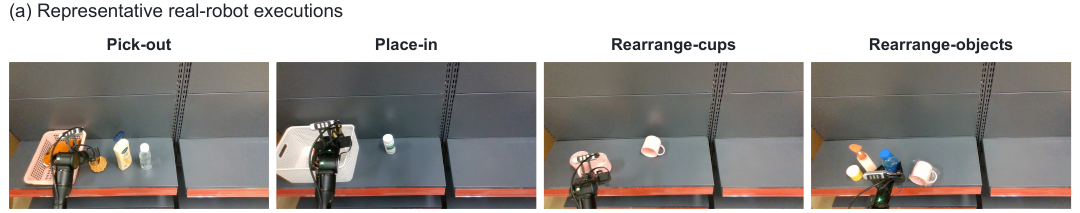}\par
\begingroup
\fontsize{8}{9.5}\selectfont
\setlength{\tabcolsep}{3pt}
\newcommand{\robotlegenditem}[2]{%
  \textcolor[HTML]{#1}{\rule{0.7em}{0.7em}}~#2}
\begin{tabular}{@{}lllllll@{}}
\robotlegenditem{969B9E}{Action-only baseline} &
\robotlegenditem{C9AA72}{LAPA} &
\robotlegenditem{DFCEA2}{LAPA-DINOv2} &
\robotlegenditem{AD8760}{LAPA-DINOv3} &
\robotlegenditem{A7C5CB}{DINOv3} &
\robotlegenditem{789EAB}{V-JEPA~2} &
\robotlegenditem{527887}{Wan2.2}
\end{tabular}\par
\endgroup
\includegraphics[width=\textwidth,trim=0 0 0 120.24bp,clip]{figures/05_supporting_evaluations_asset.pdf}
\caption{Real-robot evaluation: (a) representative executions; (b) partial-credit scores
normalized by 160 points per task (20 trials, up to 8 points each).
LA/FV use P50. Protocol: Appendix~\ref{app:beyond_standard_evaluations}.}
\label{fig:supporting_evaluations}
\end{figure}

On the four real-robot tasks, all configurations use the same training
budget, with LA and FV evaluated at P50
(Appendix~\ref{app:beyond_standard_evaluations}).
Figure~\ref{fig:supporting_evaluations} reports partial-credit
task-completion scores over 20 trials per task. Every FV configuration
scores higher than every LA configuration and the action-only baseline on
all four tasks, with Wan2.2 highest throughout, whereas no LA configuration
exceeds the action-only baseline on any task. Overall, FV thus generally
outperforms LA and, on the real robot, again surpasses action-only training
with half of its samples routed to action prediction.

\subsection{Why Can LA and FV Supervision Help Policy Learning?}
\label{sec:diagnostics}
To investigate possible explanations for these gains, we examine
action-related information recoverable from frozen source
representations, its organization in feature space, and visual attention
in trained policies.

We begin by testing what action information can be recovered from the six
frozen source representations. Separate visual-only probes predict
absolute action trajectories (ABS) or relative action changes (\(\Delta_c\))
over a 45-step segment from the frozen representations of its image pair
(for LA, the post-quantization LA vector), on CALVIN
environment-D validation and on unseen tasks of the dual-arm real-robot
dataset AgiBotWorld-Beta~\citep{agibot_world}, using LARYBench splits~\citep{rw_lary} (Appendix~\ref{app:diagnostics}). The readouts in
Table~\ref{tab:diagnostics} indicate that both LA and FV carry
action-related information, so predicting such targets may provide
indirect action supervision on auxiliary-routed samples. Their strengths differ: all three FV sources yield lower ABS
position errors than all three LA sources on both datasets, whereas both
LAPA-DINO variants yield lower relative-motion errors than all three FV
sources at every \(\Delta_c\) setting on unseen AgiBotWorld-Beta tasks; on CALVIN,
the leading encoder changes with \(c\). Plain LAPA, which
yields the weakest CALVIN readouts, is also the weakest policy
configuration on CALVIN (Table~\ref{tab:policy}).

\begin{table}[!htbp]
\centering
\small
\setlength{\tabcolsep}{4pt}
\renewcommand{\arraystretch}{1.0}
\caption{Frozen-source position errors (mm; lower is better) from visual-only
probes; $\Delta_c$ uses action differences at stride $c$ within a
45-step segment. Bold: column minima. Protocol: Appendix~\ref{app:diagnostic_protocols}.}
\label{tab:diagnostics}
\begin{tabular*}{\textwidth}{@{\extracolsep{\fill}}lrrrrrrrr@{}}
\toprule
& \multicolumn{4}{c}{AgiBotWorld-Beta unseen tasks}
& \multicolumn{4}{c}{CALVIN environment-D validation}\\
\cmidrule(lr){2-5}\cmidrule(lr){6-9}
Encoder & ABS & $\Delta_{c=1}$ & $\Delta_{c=9}$ & $\Delta_{c=45}$
& ABS & $\Delta_{c=1}$ & $\Delta_{c=9}$ & $\Delta_{c=45}$\\
\midrule
DINOv3 & \textbf{76.31} & 25.23 & 13.73 & 14.19
& \textbf{52.06} & 69.64 & 75.44 & 58.21\\
V-JEPA~2 & 101.98 & 20.44 & 11.42 & 12.16
& 54.70 & 64.21 & 68.43 & 55.85\\
Wan2.2 & 105.55 & 32.28 & 17.20 & 16.65
& 53.86 & 53.44 & 59.72 & \textbf{50.01}\\
LAPA-DINOv2 & 155.65 & 14.79 & \textbf{8.62} & \textbf{8.79}
& 62.70 & 56.89 & 59.16 & 54.40\\
LAPA-DINOv3 & 155.61 & \textbf{14.68} & 8.71 & 9.23
& 60.14 & \textbf{53.23} & \textbf{57.41} & 50.48\\
LAPA & 164.82 & 30.79 & 15.79 & 15.58
& 127.91 & 139.59 & 109.36 & 95.03\\
\bottomrule
\end{tabular*}
\end{table}

\begin{figure}[!htbp]
\centering
\includegraphics[width=\textwidth]{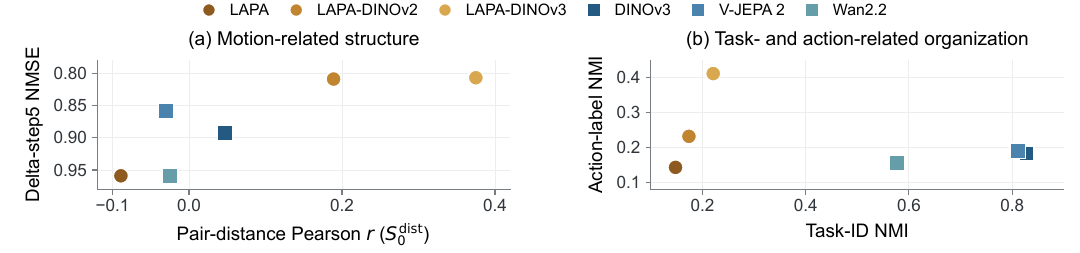}
\caption{Frozen-source diagnostics on AgiBotWorld-Beta seen validation.
(a) Correlation between net-action-change and representation pair distances, plotted against delta-step5 NMSE.
(b) Feature-cluster NMI with displacement-derived action labels and task IDs.
Each point is one encoder; details in Appendix~\ref{app:diagnostic_geometry}.}
\label{fig:clusters}
\end{figure}

We next examine how this action-related information is organized in
feature space. On AgiBotWorld-Beta seen validation (held-out segments of
the probe-training tasks), we correlate
representation distances between sample pairs with distances between
their net action changes over the segment. Figure~\ref{fig:clusters}a
plots this correlation against the normalized mean squared error (NMSE) of a second relative-motion
probe (delta-step5), which predicts the nine successive five-step action
differences within the segment. Both
LAPA-DINO variants combine higher correlations with lower readout errors
than the FV sources; LAPA-DINOv3 has the highest correlation (0.375)
and lowest NMSE (0.807). This alignment is
consistent with LAPA-DINO targets providing a learning signal that
distinguishes different movements.

Clustering on the same validation split further distinguishes
motion-related and task-related organization (Figure~\ref{fig:clusters}b).
We measure normalized mutual information (NMI) between feature-cluster
assignments and motion labels obtained by clustering ground-truth
displacements, or dataset task IDs. Both LAPA-DINO variants associate
more strongly with motion labels than any FV source, whereas all three FV
sources associate more strongly with task IDs than any LA source.
Together with the readouts, these results suggest that LAPA-DINO
supervision may emphasize motion distinctions, whereas FV supervision may
supply position- and task-associated cues. Different tasks may benefit
differently from these cues, potentially contributing to the
setting-dependent comparisons in Section~\ref{sec:supervision_characteristics}.

\begin{figure*}[!htbp]
\centering
\begingroup
\begin{minipage}[t]{0.118\textwidth}
\centering
\parbox[c][10pt][c]{\linewidth}{\centering\fontsize{7.2}{8.5}\selectfont\mbox{Input}}\par
\includegraphics[width=\linewidth]{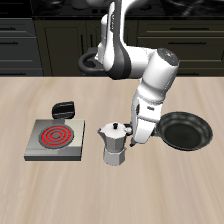}\par
\vspace{5pt}
\includegraphics[width=\linewidth]{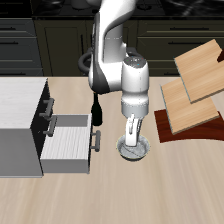}
\end{minipage}%
\hfill
\begin{minipage}[t]{0.118\textwidth}
\centering
\parbox[c][10pt][c]{\linewidth}{\centering\fontsize{7.2}{8.5}\selectfont\mbox{Action-only}}\par
\includegraphics[width=\linewidth]{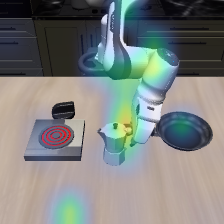}\par
\vspace{5pt}
\includegraphics[width=\linewidth]{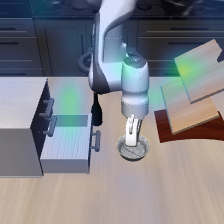}
\end{minipage}%
\hfill
\begin{minipage}[t]{0.118\textwidth}
\centering
\parbox[c][10pt][c]{\linewidth}{\centering\fontsize{7.2}{8.5}\selectfont\mbox{LAPA}}\par
\includegraphics[width=\linewidth]{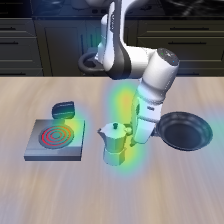}\par
\vspace{5pt}
\includegraphics[width=\linewidth]{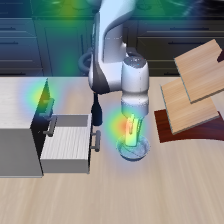}
\end{minipage}%
\hfill
\begin{minipage}[t]{0.118\textwidth}
\centering
\parbox[c][10pt][c]{\linewidth}{\centering\fontsize{7.2}{8.5}\selectfont\mbox{LAPA-DINOv2}}\par
\includegraphics[width=\linewidth]{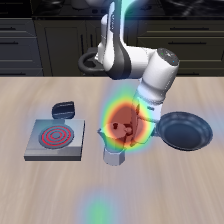}\par
\vspace{5pt}
\includegraphics[width=\linewidth]{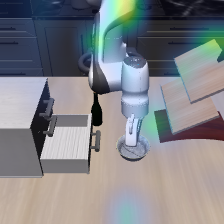}
\end{minipage}%
\hfill
\begin{minipage}[t]{0.118\textwidth}
\centering
\parbox[c][10pt][c]{\linewidth}{\centering\fontsize{7.2}{8.5}\selectfont\mbox{LAPA-DINOv3}}\par
\includegraphics[width=\linewidth]{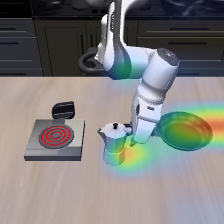}\par
\vspace{5pt}
\includegraphics[width=\linewidth]{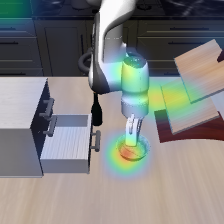}
\end{minipage}%
\hfill
\begin{minipage}[t]{0.118\textwidth}
\centering
\parbox[c][10pt][c]{\linewidth}{\centering\fontsize{7.2}{8.5}\selectfont\mbox{DINOv3}}\par
\includegraphics[width=\linewidth]{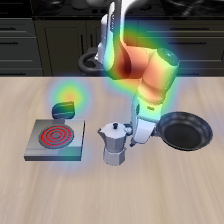}\par
\vspace{5pt}
\includegraphics[width=\linewidth]{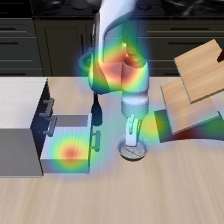}
\end{minipage}%
\hfill
\begin{minipage}[t]{0.118\textwidth}
\centering
\parbox[c][10pt][c]{\linewidth}{\centering\fontsize{7.2}{8.5}\selectfont\mbox{V-JEPA~2}}\par
\includegraphics[width=\linewidth]{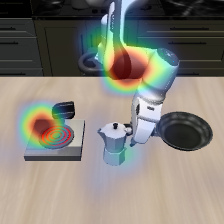}\par
\vspace{5pt}
\includegraphics[width=\linewidth]{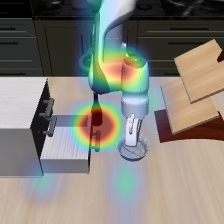}
\end{minipage}%
\hfill
\begin{minipage}[t]{0.118\textwidth}
\centering
\parbox[c][10pt][c]{\linewidth}{\centering\fontsize{7.2}{8.5}\selectfont\mbox{Wan2.2}}\par
\includegraphics[width=\linewidth]{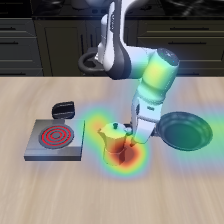}\par
\vspace{5pt}
\includegraphics[width=\linewidth]{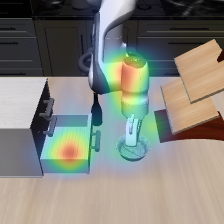}
\end{minipage}\par
\endgroup
\caption{Qualitative policy attention on two LIBERO observations, one per row.
Columns show the input observation and the action-only, LA, and FV policies.
Colors show relative intensity within each map. Both rows use the extraction
protocol in Appendix~\ref{app:policy_attention}.}
\label{fig:policy_attention}
\end{figure*}

Finally, Figure~\ref{fig:policy_attention} compares the action-only
policy and all six LA/FV configurations on two LIBERO observations
(extraction protocol in Appendix~\ref{app:policy_attention}). In the top
row, the action-only map is diffuse, whereas V-JEPA~2 FV highlights the
stove region and the robot, and LAPA-DINOv3 emphasizes areas around the
gripper and nearby containers. The bottom row shows a second observation
under the same extraction protocol. These maps suggest that LA and FV supervision can reshape the policy's
attention toward task-relevant objects and interaction regions, which may
supply more task-relevant visual information for action prediction and
contribute to the gains observed above.

\FloatBarrier

\section{Conclusion}
\label{sec:conclusion}
Robot policies must execute actions, but this does not require most
training samples to supervise direct action prediction. In a unified VLA
framework with sample-level routing, suitable LA and FV targets can outperform
action-only training under the same budget even when direct action prediction
receives no more than half of the samples, although the best allocation depends on the target,
benchmark, and budget. Between the two families, FV generally performs
better, leading on LIBERO and all four real-robot tasks, while LAPA-DINOv3
remains competitive on CALVIN. Diagnostics suggest why both can help: both
carry action-related information, with LAPA-DINO targets emphasizing
relative motion and FV targets emphasizing absolute position and task
identity, and both can redirect policy attention toward task-relevant regions.
These findings support designing predictive targets and their supervision
allocation jointly, rather than defaulting to action-only training.

\label{maintextend}

\FloatBarrier
\begingroup
\setlength{\bibsep}{5pt plus 1pt minus 1pt}
\bibliography{references}

\begin{thebibliography}{33}
\providecommand{\natexlab}[1]{#1}
\providecommand{\url}[1]{\texttt{#1}}
\expandafter\ifx\csname urlstyle\endcsname\relax
  \providecommand{\doi}[1]{doi: #1}\else
  \providecommand{\doi}{doi: \begingroup \urlstyle{rm}\Url}\fi

\bibitem[AgiBot-World-Contributors et~al.(2025)AgiBot-World-Contributors, Bu,
  Cai, Chen, Cui, Ding, Feng, Gao, He, Hu, Huang, Jiang, Jiang, Jing, Li, Li,
  Liu, Liu, Lu, Luo, Luo, Mu, Niu, Pan, Pang, Qiao, Ren, Ruan, Shan, Shen, Shi,
  Shi, Shi, Sima, Song, Wang, Wang, Wei, Xie, Xu, Yan, Yang, Yang, Yang, Yao,
  Zeng, Zhang, Zhang, Zhao, Zhao, Zhao, and Zhu]{agibot_world}
AgiBot-World-Contributors, Qingwen Bu, Jisong Cai, Li~Chen, Xiuqi Cui, Yan
  Ding, Siyuan Feng, Shenyuan Gao, Xindong He, Xuan Hu, Xu~Huang, Shu Jiang,
  Yuxin Jiang, Cheng Jing, Hongyang Li, Jialu Li, Chiming Liu, Yi~Liu, Yuxiang
  Lu, Jianlan Luo, Ping Luo, Yao Mu, Yuehan Niu, Yixuan Pan, Jiangmiao Pang,
  Yu~Qiao, Guanghui Ren, Cheng Ruan, Jiaqi Shan, Yongjian Shen, Chengshi Shi,
  Mingkang Shi, Modi Shi, Chonghao Sima, Jianheng Song, Huijie Wang, Wenhao
  Wang, Dafeng Wei, Chengen Xie, Guo Xu, Junchi Yan, Cunbiao Yang, Lei Yang,
  Shukai Yang, Maoqing Yao, Jia Zeng, Chi Zhang, Qinglin Zhang, Bin Zhao,
  Chengyue Zhao, Jiaqi Zhao, and Jianchao Zhu.
\newblock {AgiBot World Colosseo}: A large-scale manipulation platform for
  scalable and intelligent embodied systems.
\newblock \emph{arXiv preprint arXiv:2503.06669}, 2025.
\newblock URL \url{https://arxiv.org/abs/2503.06669}.

\bibitem[Assran et~al.(2025)Assran, Bardes, Fan, Garrido, Howes, Komeili,
  Muckley, Rizvi, Roberts, Sinha, Zholus, Arnaud, Gejji, Martin, Hogan, Dugas,
  Bojanowski, Khalidov, Labatut, Massa, Szafraniec, Krishnakumar, Li, Ma,
  Chandar, Meier, LeCun, Rabbat, and Ballas]{vjepa2}
Mido Assran, Adrien Bardes, David Fan, Quentin Garrido, Russell Howes, Mojtaba
  Komeili, Matthew Muckley, Ammar Rizvi, Claire Roberts, Koustuv Sinha, Artem
  Zholus, Sergio Arnaud, Abha Gejji, Ada Martin, Francois~Robert Hogan, Daniel
  Dugas, Piotr Bojanowski, Vasil Khalidov, Patrick Labatut, Francisco Massa,
  Marc Szafraniec, Kapil Krishnakumar, Yong Li, Xiaodong Ma, Sarath Chandar,
  Franziska Meier, Yann LeCun, Michael Rabbat, and Nicolas Ballas.
\newblock {V-JEPA 2}: Self-supervised video models enable understanding,
  prediction and planning.
\newblock \emph{arXiv preprint arXiv:2506.09985}, 2025.
\newblock URL \url{https://arxiv.org/abs/2506.09985}.

\bibitem[Black et~al.(2025{\natexlab{a}})Black, Brown, Darpinian, Dhabalia,
  Driess, Esmail, Equi, Finn, Fusai, Galliker, Ghosh, Groom, Hausman, Ichter,
  Jakubczak, Jones, Ke, LeBlanc, Levine, Li-Bell, Mothukuri, Nair, Pertsch,
  Ren, Shi, Smith, Springenberg, Stachowicz, Tanner, Vuong, Walke, Walling,
  Wang, Yu, and Zhilinsky]{baseline_pi05}
Kevin Black, Noah Brown, James Darpinian, Karan Dhabalia, Danny Driess, Adnan
  Esmail, Michael~Robert Equi, Chelsea Finn, Niccolo Fusai, Manuel~Y. Galliker,
  Dibya Ghosh, Lachy Groom, Karol Hausman, Brian Ichter, Szymon Jakubczak, Tim
  Jones, Liyiming Ke, Devin LeBlanc, Sergey Levine, Adrian Li-Bell, Mohith
  Mothukuri, Suraj Nair, Karl Pertsch, Allen~Z. Ren, Lucy~Xiaoyang Shi, Laura
  Smith, Jost~Tobias Springenberg, Kyle Stachowicz, James Tanner, Quan Vuong,
  Homer Walke, Anna Walling, Haohuan Wang, Lili Yu, and Ury Zhilinsky.
\newblock {$\pi_{0.5}$}: a vision-language-action model with open-world
  generalization.
\newblock In Joseph Lim, Shuran Song, and Hae-Won Park (eds.),
  \emph{Proceedings of The 9th Conference on Robot Learning}, volume 305 of
  \emph{Proceedings of Machine Learning Research}, pp.\  17--40. PMLR, 27--30
  Sep 2025{\natexlab{a}}.
\newblock URL \url{https://proceedings.mlr.press/v305/black25a.html}.

\bibitem[Black et~al.(2025{\natexlab{b}})Black, Brown, Driess, Esmail, Equi,
  Finn, Fusai, Groom, Hausman, Ichter, Jakubczak, Jones, Ke, Levine, Li-Bell,
  Mothukuri, Nair, Pertsch, Shi, Smith, Tanner, Vuong, Walling, Wang, and
  Zhilinsky]{baseline_pi0}
Kevin Black, Noah Brown, Danny Driess, Adnan Esmail, Michael~Robert Equi,
  Chelsea Finn, Niccolo Fusai, Lachy Groom, Karol Hausman, Brian Ichter, Szymon
  Jakubczak, Tim Jones, Liyiming Ke, Sergey Levine, Adrian Li-Bell, Mohith
  Mothukuri, Suraj Nair, Karl Pertsch, Lucy~Xiaoyang Shi, Laura Smith, James
  Tanner, Quan Vuong, Anna Walling, Haohuan Wang, and Ury Zhilinsky.
\newblock {$\pi_0$}: A vision-language-action flow model for general robot
  control.
\newblock In \emph{Proceedings of Robotics: Science and Systems}, Los Angeles,
  CA, USA, June 2025{\natexlab{b}}.
\newblock \doi{10.15607/RSS.2025.XXI.010}.

\bibitem[Bu et~al.(2025)Bu, Yang, Cai, Gao, Ren, Yao, Luo, and Li]{rw_univla}
Qingwen Bu, Yanting Yang, Jisong Cai, Shenyuan Gao, Guanghui Ren, Maoqing Yao,
  Ping Luo, and Hongyang Li.
\newblock Learning to act anywhere with task-centric latent actions.
\newblock In \emph{Proceedings of Robotics: Science and Systems}, Los Angeles,
  CA, USA, June 2025.
\newblock \doi{10.15607/RSS.2025.XXI.014}.

\bibitem[Bu et~al.(2026)Bu, Hu, Zhou, Zhang, Tang, Liu, Tao, Ma, Huang, Tang,
  Wang, Zhang, Ma, Ye, Li, and Hao]{intro_what_matters}
Xizhou Bu, Qingda Hu, Lei Zhou, Lingfeng Zhang, Yingbo Tang, Zihao Liu, Xinyi
  Tao, Zhiqiang Ma, Qingqiu Huang, Chufeng Tang, Hongbo Wang, Jing Zhang, Jiayi
  Ma, Hangjun Ye, Wei Li, and Xiaoshuai Hao.
\newblock What matters for latent actions in robot learning.
\newblock \emph{arXiv preprint arXiv:2608.19613}, 2026.
\newblock URL \url{https://arxiv.org/abs/2608.19613}.

\bibitem[Cheang et~al.(2024)Cheang, Chen, Jing, Kong, Li, Li, Liu, Wu, Xu,
  Yang, Zhang, and Zhu]{rw_gr2}
Chi-Lam Cheang, Guangzeng Chen, Ya~Jing, Tao Kong, Hang Li, Yifeng Li, Yuxiao
  Liu, Hongtao Wu, Jiafeng Xu, Yichu Yang, Hanbo Zhang, and Minzhao Zhu.
\newblock {GR-2}: A generative video-language-action model with web-scale
  knowledge for robot manipulation.
\newblock \emph{arXiv preprint arXiv:2410.06158}, 2024.
\newblock URL \url{https://arxiv.org/abs/2410.06158}.

\bibitem[Chen et~al.(2026)Chen, Wei, Zhang, Zhang, Wang, Guo, Yang, Wang, Xiao,
  Zhao, Chen, and Bian]{intro_villax}
Xiaoyu Chen, Hangxing Wei, Pushi Zhang, Chuheng Zhang, Kaixin Wang, Yanjiang
  Guo, Rushuai Yang, Yucen Wang, Xinquan Xiao, Li~Zhao, Jianyu Chen, and Jiang
  Bian.
\newblock {villa-X}: Enhancing latent action modeling in vision-language-action
  models.
\newblock In \emph{International Conference on Learning Representations}, 2026.
\newblock URL
  \url{https://proceedings.iclr.cc/paper_files/paper/2026/hash/73471f46f69a13cede6cf2db3a6f438f-Abstract-Conference.html}.

\bibitem[Guo et~al.(2025)Guo, Cao, Tao, Xu, Yan, Liang, Laptev, and
  Chang]{intro_glad}
Minghao Guo, Meng Cao, Jiachen Tao, Rongtao Xu, Yan Yan, Xiaodan Liang, Ivan
  Laptev, and Xiaojun Chang.
\newblock {GLaD}: Geometric latent distillation for vision-language-action
  models.
\newblock \emph{arXiv preprint arXiv:2512.09619}, 2025.
\newblock URL \url{https://arxiv.org/abs/2512.09619}.

\bibitem[Hu et~al.(2025)Hu, Guo, Wang, Chen, Wang, Zhang, Sreenath, Lu, and
  Chen]{intro_vpp}
Yucheng Hu, Yanjiang Guo, Pengchao Wang, Xiaoyu Chen, Yen-Jen Wang, Jianke
  Zhang, Koushil Sreenath, Chaochao Lu, and Jianyu Chen.
\newblock Video prediction policy: A generalist robot policy with predictive
  visual representations.
\newblock In Aarti Singh, Maryam Fazel, Daniel Hsu, Simon Lacoste-Julien, Felix
  Berkenkamp, Tegan Maharaj, Kiri Wagstaff, and Jerry Zhu (eds.),
  \emph{Proceedings of the 42nd International Conference on Machine Learning},
  volume 267 of \emph{Proceedings of Machine Learning Research}, pp.\
  24328--24346. PMLR, 13--19 Jul 2025.
\newblock URL \url{https://proceedings.mlr.press/v267/hu25g.html}.

\bibitem[Karamcheti et~al.(2023)Karamcheti, Nair, Chen, Kollar, Finn, Sadigh,
  and Liang]{rw_voltron}
Siddharth Karamcheti, Suraj Nair, Annie~S Chen, Thomas Kollar, Chelsea Finn,
  Dorsa Sadigh, and Percy Liang.
\newblock Language-driven representation learning for robotics.
\newblock In \emph{Proceedings of Robotics: Science and Systems}, Daegu,
  Republic of Korea, July 2023.
\newblock \doi{10.15607/RSS.2023.XIX.032}.

\bibitem[Kim et~al.(2025)Kim, Pertsch, Karamcheti, Xiao, Balakrishna, Nair,
  Rafailov, Foster, Sanketi, Vuong, Kollar, Burchfiel, Tedrake, Sadigh, Levine,
  Liang, and Finn]{baseline_openvla}
Moo~Jin Kim, Karl Pertsch, Siddharth Karamcheti, Ted Xiao, Ashwin Balakrishna,
  Suraj Nair, Rafael Rafailov, Ethan~P Foster, Pannag~R Sanketi, Quan Vuong,
  Thomas Kollar, Benjamin Burchfiel, Russ Tedrake, Dorsa Sadigh, Sergey Levine,
  Percy Liang, and Chelsea Finn.
\newblock {OpenVLA}: An open-source vision-language-action model.
\newblock In Pulkit Agrawal, Oliver Kroemer, and Wolfram Burgard (eds.),
  \emph{Proceedings of The 8th Conference on Robot Learning}, volume 270 of
  \emph{Proceedings of Machine Learning Research}, pp.\  2679--2713. PMLR,
  06--09 Nov 2025.
\newblock URL \url{https://proceedings.mlr.press/v270/kim25c.html}.

\bibitem[Li et~al.(2026)Li, Song, Zhao, Wang, Ding, Wang, Zeng, and
  Li]{intro_spatial_forcing}
Fuhao Li, Wenxuan Song, Han Zhao, Jingbo Wang, Pengxiang Ding, Donglin Wang,
  Long Zeng, and Haoang Li.
\newblock {Spatial Forcing}: Implicit spatial representation alignment for
  vision-language-action model.
\newblock In \emph{International Conference on Learning Representations}, pp.\
  132324--132345, 2026.
\newblock URL
  \url{https://proceedings.iclr.cc/paper_files/paper/2026/hash/d676c05a8fb93ad9231508d95567c498-Abstract-Conference.html}.

\bibitem[Lin et~al.(2026)Lin, Li, Li, Shen, Zhao, Shao, and
  Zhang]{intro_pixels_to_tokens}
Yihan Lin, Haoyang Li, Yang Li, Haitao Shen, Yihan Zhao, Chao Shao, and Jing
  Zhang.
\newblock From pixels to tokens: A systematic study of latent action
  supervision for vision-language-action models.
\newblock In \emph{Forty-third International Conference on Machine Learning},
  2026.
\newblock URL \url{https://openreview.net/forum?id=VMsumctGvg}.

\bibitem[Liu et~al.(2023)Liu, Zhu, Gao, Feng, Liu, Zhu, and
  Stone]{benchmark_libero}
Bo~Liu, Yifeng Zhu, Chongkai Gao, Yihao Feng, Qiang Liu, Yuke Zhu, and Peter
  Stone.
\newblock {LIBERO}: Benchmarking knowledge transfer for lifelong robot
  learning.
\newblock In \emph{Advances in Neural Information Processing Systems},
  volume~36, 2023.
\newblock URL
  \url{https://proceedings.neurips.cc/paper_files/paper/2023/hash/8c3c666820ea055a77726d66fc7d447f-Abstract-Datasets_and_Benchmarks.html}.

\bibitem[Mees et~al.(2022)Mees, Hermann, Rosete-Beas, and
  Burgard]{benchmark_calvin}
Oier Mees, Lukas Hermann, Erick Rosete-Beas, and Wolfram Burgard.
\newblock {CALVIN}: A benchmark for language-conditioned policy learning for
  long-horizon robot manipulation tasks.
\newblock \emph{IEEE Robotics and Automation Letters}, 7\penalty0 (3):\penalty0
  7327--7334, 2022.
\newblock \doi{10.1109/LRA.2022.3180108}.

\bibitem[{Meituan LongCat}(2026)]{larybench_repo}
{Meituan LongCat}.
\newblock {LARYBench} official repository.
\newblock GitHub repository, 2026.
\newblock URL \url{https://github.com/meituan-longcat/LARYBench}.

\bibitem[Nie et~al.(2026)Nie, Chen, Lv, Kuang, Li, Cao, and Cai]{rw_lary}
Dujun Nie, Fengjiao Chen, Qi~Lv, Jun Kuang, Xiaoyu Li, Xuezhi Cao, and Xunliang
  Cai.
\newblock {LARY}: A latent action representation yielding benchmark for
  generalizable vision-to-action alignment.
\newblock \emph{arXiv preprint arXiv:2604.11689}, 2026.
\newblock URL \url{https://arxiv.org/abs/2604.11689}.

\bibitem[{OpenBMB}(2026)]{minicpmv46_modelcard}
{OpenBMB}.
\newblock {MiniCPM-V-4.6} model card.
\newblock Hugging Face model repository, 2026.
\newblock URL \url{https://huggingface.co/openbmb/MiniCPM-V-4.6}.

\bibitem[Oquab et~al.(2024)Oquab, Darcet, Moutakanni, Vo, Szafraniec, Khalidov,
  Fernandez, Haziza, Massa, El-Nouby, Assran, Ballas, Galuba, Howes, Huang, Li,
  Misra, Rabbat, Sharma, Synnaeve, Xu, Jegou, Mairal, Labatut, Joulin, and
  Bojanowski]{dinov2}
Maxime Oquab, Timoth{\'e}e Darcet, Th{\'e}o Moutakanni, Huy~V. Vo, Marc
  Szafraniec, Vasil Khalidov, Pierre Fernandez, Daniel Haziza, Francisco Massa,
  Alaaeldin El-Nouby, Mido Assran, Nicolas Ballas, Wojciech Galuba, Russell
  Howes, Po-Yao Huang, Shang-Wen Li, Ishan Misra, Michael Rabbat, Vasu Sharma,
  Gabriel Synnaeve, Hu~Xu, Herve Jegou, Julien Mairal, Patrick Labatut, Armand
  Joulin, and Piotr Bojanowski.
\newblock {DINO}v2: Learning robust visual features without supervision.
\newblock \emph{Transactions on Machine Learning Research}, 2024.
\newblock ISSN 2835-8856.
\newblock URL \url{https://openreview.net/forum?id=a68SUt6zFt}.
\newblock Featured Certification.

\bibitem[Sim{\'e}oni et~al.(2026)Sim{\'e}oni, Vo, Seitzer, Baldassarre, Oquab,
  Jose, Khalidov, Szafraniec, Yi, Ramamonjisoa, Massa, Haziza, Wehrstedt, Wang,
  Darcet, Moutakanni, Sentana, Roberts, Vedaldi, Tolan, Brandt, Couprie,
  Mairal, Jegou, Labatut, and Bojanowski]{dinov3}
Oriane Sim{\'e}oni, Huy~V. Vo, Maximilian Seitzer, Federico Baldassarre, Maxime
  Oquab, Cijo Jose, Vasil Khalidov, Marc Szafraniec, Seung~Eun Yi, Michael
  Ramamonjisoa, Francisco Massa, Daniel Haziza, Luca Wehrstedt, Jianyuan Wang,
  Timoth{\'e}e Darcet, Th{\'e}o Moutakanni, Leonel Sentana, Claire Roberts,
  Andrea Vedaldi, Jamie Tolan, John Brandt, Camille Couprie, Julien Mairal,
  Herve Jegou, Patrick Labatut, and Piotr Bojanowski.
\newblock {DINO}v3.
\newblock \emph{Transactions on Machine Learning Research}, 2026.
\newblock ISSN 2835-8856.
\newblock URL \url{https://openreview.net/forum?id=2NlGyqNjns}.
\newblock Featured Certification.

\bibitem[Tian et~al.(2025)Tian, Yang, Zeng, Wang, Lin, Dong, and Pang]{rw_seer}
Yang Tian, Sizhe Yang, Jia Zeng, Ping Wang, Dahua Lin, Hao Dong, and Jiangmiao
  Pang.
\newblock Predictive inverse dynamics models are scalable learners for robotic
  manipulation.
\newblock In \emph{International Conference on Learning Representations}, 2025.
\newblock URL
  \url{https://proceedings.iclr.cc/paper_files/paper/2025/hash/e5b5c402bb7bd5e60bede6961d6fe39e-Abstract-Conference.html}.

\bibitem[Wan et~al.(2025)Wan, Wang, Ai, Wen, Mao, Xie, Chen, Yu, Zhao, Yang,
  Zeng, Wang, Zhang, Zhou, Wang, Chen, Zhu, Zhao, Yan, Huang, Feng, Zhang, Li,
  Wu, Chu, Feng, Zhang, Sun, Fang, Wang, Gui, Weng, Shen, Lin, Wang, Wang,
  Zhou, Wang, Shen, Yu, Shi, Huang, Xu, Kou, Lv, Li, Liu, Wang, Zhang, Huang,
  Li, Wu, Liu, Pan, Zheng, Hong, Shi, Feng, Jiang, Han, Wu, and Liu]{wan}
Team Wan, Ang Wang, Baole Ai, Bin Wen, Chaojie Mao, Chen-Wei Xie, Di~Chen,
  Feiwu Yu, Haiming Zhao, Jianxiao Yang, Jianyuan Zeng, Jiayu Wang, Jingfeng
  Zhang, Jingren Zhou, Jinkai Wang, Jixuan Chen, Kai Zhu, Kang Zhao, Keyu Yan,
  Lianghua Huang, Mengyang Feng, Ningyi Zhang, Pandeng Li, Pingyu Wu, Ruihang
  Chu, Ruili Feng, Shiwei Zhang, Siyang Sun, Tao Fang, Tianxing Wang, Tianyi
  Gui, Tingyu Weng, Tong Shen, Wei Lin, Wei Wang, Wei Wang, Wenmeng Zhou, Wente
  Wang, Wenting Shen, Wenyuan Yu, Xianzhong Shi, Xiaoming Huang, Xin Xu, Yan
  Kou, Yangyu Lv, Yifei Li, Yijing Liu, Yiming Wang, Yingya Zhang, Yitong
  Huang, Yong Li, You Wu, Yu~Liu, Yulin Pan, Yun Zheng, Yuntao Hong, Yupeng
  Shi, Yutong Feng, Zeyinzi Jiang, Zhen Han, Zhi-Fan Wu, and Ziyu Liu.
\newblock {Wan}: Open and advanced large-scale video generative models.
\newblock \emph{arXiv preprint arXiv:2503.20314}, 2025.
\newblock URL \url{https://arxiv.org/abs/2503.20314}.

\bibitem[{Wan Team}(2025)]{wan22_repo}
{Wan Team}.
\newblock {Wan2.2} official repository.
\newblock GitHub repository, 2025.
\newblock URL \url{https://github.com/Wan-Video/Wan2.2}.

\bibitem[Wu et~al.(2024)Wu, Jing, Cheang, Chen, Xu, Li, Liu, Li, and
  Kong]{intro_gr1}
Hongtao Wu, Ya~Jing, Chilam Cheang, Guangzeng Chen, Jiafeng Xu, Xinghang Li,
  Minghuan Liu, Hang Li, and Tao Kong.
\newblock Unleashing large-scale video generative pre-training for visual robot
  manipulation.
\newblock In \emph{International Conference on Learning Representations}, 2024.
\newblock URL
  \url{https://proceedings.iclr.cc/paper_files/paper/2024/hash/2c37c5bcef24b9541550261dcd63261b-Abstract-Conference.html}.

\bibitem[Yang et~al.(2026)Yang, Song, Wang, Sheng, Fang, Zhou, He, Yan, Chen,
  Sun, Sun, Wang, Liu, Wang, Gao, Dayoub, and Li]{rw_4d_wam}
Lishan Yang, Wenxuan Song, Xi~Wang, Pingyue Sheng, Zheng Fang, Ziyang Zhou,
  Junjie He, Haodong Yan, Jiayi Chen, Nan Sun, Qiao Sun, Pengwei Wang, Lingqiao
  Liu, Yan Wang, Yuxiang Gao, Feras Dayoub, and Haoang Li.
\newblock {4D-WAM}: Infusing spatiotemporal awareness into world action models
  through trajectory fields.
\newblock \emph{arXiv preprint arXiv:2608.08023}, 2026.
\newblock URL \url{https://arxiv.org/abs/2608.08023}.

\bibitem[Yao et~al.(2025)Yao, Yu, Zhang, Wang, Cui, Zhu, Cai, Chen, Li, Zhao,
  He, Chen, Zhou, Zou, Zhang, Hu, Zheng, Zhou, Cai, Han, Zeng, Li, Liu, and
  Sun]{minicpmv_series}
Yuan Yao, Tianyu Yu, Ao~Zhang, Chongyi Wang, Junbo Cui, Hongji Zhu, Tianchi
  Cai, Chi Chen, Haoyu Li, Weilin Zhao, Zhihui He, Qianyu Chen, Ronghua Zhou,
  Zhensheng Zou, Haoye Zhang, Shengding Hu, Zhi Zheng, Jie Zhou, Jie Cai,
  Xu~Han, Guoyang Zeng, Dahai Li, Zhiyuan Liu, and Maosong Sun.
\newblock Efficient {GPT-4V} level multimodal large language model for
  deployment on edge devices.
\newblock \emph{Nature Communications}, 16:\penalty0 5509, 2025.
\newblock \doi{10.1038/s41467-025-61040-5}.
\newblock URL \url{https://www.nature.com/articles/s41467-025-61040-5}.

\bibitem[Ye et~al.(2025)Ye, Jang, Jeon, Joo, Yang, Peng, Mandlekar, Tan, Chao,
  Lin, Liden, Lee, Gao, Zettlemoyer, Fox, and Seo]{intro_lapa}
Seonghyeon Ye, Joel Jang, Byeongguk Jeon, Se~June Joo, Jianwei Yang, Baolin
  Peng, Ajay Mandlekar, Reuben Tan, Yu-Wei Chao, Bill~Yuchen Lin, Lars Liden,
  Kimin Lee, Jianfeng Gao, Luke Zettlemoyer, Dieter Fox, and Minjoon Seo.
\newblock Latent action pretraining from videos.
\newblock In \emph{International Conference on Learning Representations}, 2025.
\newblock URL
  \url{https://proceedings.iclr.cc/paper_files/paper/2025/hash/45d74e190008c7bff2845ffc8e3facd3-Abstract-Conference.html}.

\bibitem[Ye et~al.(2026)Ye, Ge, Zheng, Gao, Yu, Kurian, Indupuru, Tan, Zhu,
  Xiang, Malik, Lee, Liang, Ranawaka, Gu, Xu, Wang, Hu, Narayan, Bjorck, Wang,
  Kim, Niu, Zheng, Xie, Wu, Wang, Julian, Xu, Du, Chebotar, Reed, Kautz, Zhu,
  Fan, and Jang]{rw_dreamzero}
Seonghyeon Ye, Yunhao Ge, Kaiyuan Zheng, Shenyuan Gao, Sihyun Yu, George
  Kurian, Suneel Indupuru, You~Liang Tan, Chuning Zhu, Jiannan Xiang, Ayaan
  Malik, Kyungmin Lee, William Liang, Nadun Ranawaka, Jiasheng Gu, Yinzhen Xu,
  Guanzhi Wang, Fengyuan Hu, Avnish Narayan, Johan Bjorck, Jing Wang, Gwanghyun
  Kim, Dantong Niu, Ruijie Zheng, Yuqi Xie, Jimmy Wu, Qi~Wang, Ryan Julian,
  Danfei Xu, Yilun Du, Yevgen Chebotar, Scott Reed, Jan Kautz, Yuke Zhu,
  Linxi~"Jim" Fan, and Joel Jang.
\newblock World action models are zero-shot policies.
\newblock \emph{arXiv preprint arXiv:2602.15922}, 2026.
\newblock URL \url{https://arxiv.org/abs/2602.15922}.

\bibitem[Zhang et~al.(2025)Zhang, Pearce, Zhang, Wang, Chen, Shen, Zhao, and
  Bian]{rw_lam_learn}
Chuheng Zhang, Tim Pearce, Pushi Zhang, Kaixin Wang, Xiaoyu Chen, Wei Shen,
  Li~Zhao, and Jiang Bian.
\newblock What do latent action models actually learn?
\newblock In \emph{Advances in Neural Information Processing Systems},
  volume~38, 2025.
\newblock URL
  \url{https://proceedings.neurips.cc/paper_files/paper/2025/hash/d7b3cef7c31b94a4a533db83d01a8882-Abstract-Conference.html}.

\bibitem[Zheng et~al.(2025)Zheng, Wang, Reed, Bjorck, Fang, Hu, Jang, Kundalia,
  Lin, Magne, Narayan, Tan, Wang, Wang, Xiang, Xu, Ye, Kautz, Huang, Zhu, and
  Fan]{rw_flare}
Ruijie Zheng, Jing Wang, Scott Reed, Johan Bjorck, Yu~Fang, Fengyuan Hu, Joel
  Jang, Kaushil Kundalia, Zongyu Lin, Lo\"{i}c Magne, Avnish Narayan, You~Liang
  Tan, Guanzhi Wang, Qi~Wang, Jiannan Xiang, Yinzhen Xu, Seonghyeon Ye, Jan
  Kautz, Furong Huang, Yuke Zhu, and Linxi Fan.
\newblock {FLARE}: Robot learning with implicit world modeling.
\newblock In Joseph Lim, Shuran Song, and Hae-Won Park (eds.),
  \emph{Proceedings of The 9th Conference on Robot Learning}, volume 305 of
  \emph{Proceedings of Machine Learning Research}, pp.\  3952--3971. PMLR,
  27--30 Sep 2025.
\newblock URL \url{https://proceedings.mlr.press/v305/zheng25a.html}.

\bibitem[Zhu et~al.(2025)Zhu, Yu, Feng, Burchfiel, Shah, and Gupta]{rw_uwm}
Chuning Zhu, Raymond Yu, Siyuan Feng, Benjamin Burchfiel, Paarth Shah, and
  Abhishek Gupta.
\newblock Unified world models: Coupling video and action diffusion for
  pretraining on large robotic datasets.
\newblock In \emph{Proceedings of Robotics: Science and Systems}, Los Angeles,
  CA, USA, June 2025.
\newblock \doi{10.15607/RSS.2025.XXI.015}.

\bibitem[Zitkovich et~al.(2023)Zitkovich, Yu, Xu, Xu, Xiao, Xia, Wu, Wohlhart,
  Welker, Wahid, Vuong, Vanhoucke, Tran, Soricut, Singh, Singh, Sermanet,
  Sanketi, Salazar, Ryoo, Reymann, Rao, Pertsch, Mordatch, Michalewski, Lu,
  Levine, Lee, Lee, Leal, Kuang, Kalashnikov, Julian, Joshi, Irpan, Ichter,
  Hsu, Herzog, Hausman, Gopalakrishnan, Fu, Florence, Finn, Dubey, Driess,
  Ding, Choromanski, Chen, Chebotar, Carbajal, Brown, Brohan, Arenas, and
  Han]{rw_rt2}
Brianna Zitkovich, Tianhe Yu, Sichun Xu, Peng Xu, Ted Xiao, Fei Xia, Jialin Wu,
  Paul Wohlhart, Stefan Welker, Ayzaan Wahid, Quan Vuong, Vincent Vanhoucke,
  Huong Tran, Radu Soricut, Anikait Singh, Jaspiar Singh, Pierre Sermanet,
  Pannag~R. Sanketi, Grecia Salazar, Michael~S. Ryoo, Krista Reymann, Kanishka
  Rao, Karl Pertsch, Igor Mordatch, Henryk Michalewski, Yao Lu, Sergey Levine,
  Lisa Lee, Tsang-Wei~Edward Lee, Isabel Leal, Yuheng Kuang, Dmitry
  Kalashnikov, Ryan Julian, Nikhil~J. Joshi, Alex Irpan, Brian Ichter, Jasmine
  Hsu, Alexander Herzog, Karol Hausman, Keerthana Gopalakrishnan, Chuyuan Fu,
  Pete Florence, Chelsea Finn, Kumar~Avinava Dubey, Danny Driess, Tianli Ding,
  Krzysztof~Marcin Choromanski, Xi~Chen, Yevgen Chebotar, Justice Carbajal,
  Noah Brown, Anthony Brohan, Montserrat~Gonzalez Arenas, and Kehang Han.
\newblock {RT-2}: Vision-language-action models transfer web knowledge to
  robotic control.
\newblock In \emph{Proceedings of The 7th Conference on Robot Learning}, volume
  229 of \emph{Proceedings of Machine Learning Research}, pp.\  2165--2183.
  PMLR, 2023.
\newblock URL \url{https://proceedings.mlr.press/v229/zitkovich23a.html}.

\end{thebibliography}
\bibliographystyle{references}
\endgroup

\ifincludeappendix
  \clearpage
  \appendix
\section{Training and Evaluation Details}
\label{app:implementation}

\subsection{Target Encoders and Label Construction}
\label{app:available_configuration}
Table~\ref{tab:implementation_records} records the policy VLM and the
frozen target encoders used in the reported experiments. The frozen-source
diagnostics in Section~\ref{sec:diagnostics} use the same six target
encoders and checkpoints.

\begin{table}[!htb]
\caption{Backbones, checkpoints, and input configurations used in the
reported experiments. Parameter counts describe the loaded model or encoder.
LA predicts stored integer codes; FV predicts a velocity target constructed
from continuous encoder outputs.}
\label{tab:implementation_records}
\centering
\begin{tabularx}{\linewidth}{@{}>{\raggedright\arraybackslash}p{.17\linewidth}>{\raggedright\arraybackslash}p{.29\linewidth}>{\raggedright\arraybackslash}X@{}}
\toprule
Component & Provider and checkpoint & Configuration used\\
\midrule
Policy VLM & OpenBMB; \nolinkurl{openbmb/MiniCPM-V-4.6}
\citep{minicpmv_series,minicpmv46_modelcard} & Qwen3.5-0.8B language model
and SigLIP2-400M vision encoder; approximately 1.30B locally loaded
parameters. The text tower has hidden size 1024 and 24 layers; the vision
tower has hidden size 1152 and 27 layers. The processor uses a base
resolution of 448 and supports dynamic slicing.\\
LAPA LA & Released LARYBench asset \texttt{laq\_openx.pt}
\citep{rw_lary,larybench_repo} & Image-based latent-action quantizer; four codes per
transition and codebook size eight.\\
LAPA-DINOv2 LA & Meta FAIR; \nolinkurl{facebook/dinov2-large}
\citep{dinov2} plus \texttt{laq\_dinov2.pt} \citep{rw_lary,larybench_repo} &
ViT-L/14, approximately 304M parameters; 24 layers, 16 heads, hidden size
1024. The latent-action branch uses $224\times224$ inputs, producing a
$16\times16$ final-layer patch grid and 16 codes per transition.\\
LAPA-DINOv3 LA & Meta FAIR;
\texttt{facebook/dinov3-}\allowbreak\texttt{vitl16-pretrain-}\allowbreak
\texttt{lvd1689m} \citep{dinov3} plus
\texttt{laq\_dinov3.pt} \citep{rw_lary,larybench_repo} & ViT-L/16, approximately 303M
parameters; 24 layers, 16 heads, hidden size 1024. The latent-action branch
uses $224\times224$ inputs, producing a $14\times14$ final-layer patch grid
after excluding CLS and four register tokens, and 16 codes per transition.\\
DINOv3 FV & Meta FAIR;
\texttt{facebook/dinov3-}\allowbreak\texttt{vits16-pretrain-}\allowbreak
\texttt{lvd1689m} \citep{dinov3} &
ViT-S/16, approximately 21.6M parameters; 12 layers, six heads, hidden size
384. The FV branch uses $224\times224$ inputs and final-layer continuous
features.\\
V-JEPA~2 FV & Meta FAIR; official V-JEPA~2 release,
\texttt{vjepa2\_vitl16.pt} \citep{vjepa2} & ViT-L/16 target encoder
(the predictor is not loaded), approximately 304M parameters; 24 layers,
16 heads, hidden size 1024, temporal tubelet size two, and rotary position
embeddings.\\
Wan2.2 FV & Alibaba/Wan Team; official Wan2.2 repository
\citep{wan,wan22_repo} & Wan2.2 VAE encoder with 48 latent channels
and temporal/spatial compression $4\times16\times16$. The FV branch
encodes frames separately, mapping $224\times224$ inputs to
$14\times14\times48$ latents.\\
\bottomrule
\end{tabularx}
\end{table}

\paragraph{DINO backbones and latent-action quantizers.}
Each LAPA-DINO variant pairs the listed DINO backbone with a separately
trained LARYBench quantizer, \texttt{laq\_dinov2.pt} or
\texttt{laq\_dinov3.pt} \citep{rw_lary,larybench_repo}; both quantizers use
32-dimensional quantization with codebook size eight. The DINOv3 FV encoder
is the separate ViT-S/16 checkpoint in
Table~\ref{tab:implementation_records}.

\paragraph{LA target layout.}
With an action horizon of $H=8$, an LA target
contains eight code groups:
\begin{equation}
 z_t=(c_{t+j})_{j=0}^{7},\qquad
 c_u=(c_{u,1},\ldots,c_{u,K}).
\label{eq:la_targets}
\end{equation}
LAPA provides $K=4$ indices per group, while the two LAPA-DINO
variants provide $K=16$. Flattening the eight groups gives the
$L_z=HK$ target positions and learnable LA queries used in
Section~\ref{sec:two_stream_policy}. These are categorical indices rather
than continuous codebook vectors.

\paragraph{LA temporal construction.}
Both CALVIN and LIBERO use primary-camera transitions eight frames apart
($\delta=8$).
Let $I_u$ denote the primary-camera frame in observation $o_u$.
A code group computed from $(I_u,I_{u+8})$ is stored at source index $u$.
For a training anchor $t$ with horizon $H=8$, it retrieves groups for
$u=t,\ldots,t+7$:
\begin{equation}
 c_{t+j}=E_{\mathrm{LA}}^{(m)}(I_{t+j},I_{t+j+8}),
 \qquad j=0,\ldots,7.
\label{eq:la_temporal_construction}
\end{equation}
The final group therefore depends on $I_{t+15}$.
On LIBERO, these code groups are precomputed offline with the same
construction and read together with the actions at $t,\ldots,t+7$.
For FV supervision with horizon eight, the future observation is taken
at $t+8$. Temporal validity masks exclude anchors without the observations
or labels required by the selected target.

\paragraph{Initialization and frozen components.}
The VLM is initialized from pretrained MiniCPM-V-4.6 and fine-tuned in
full with FP32 parameters; the MMDiT prediction head is randomly
initialized. All FV encoders run frozen and without gradients, while the
projectors that map their features into the MMDiT remain trainable.
\FloatBarrier

\subsection{Implementation, Optimization, and Evaluation}
\label{app:protocol_records}
\paragraph{Noise-time sampling.}
The trajectory index $t$ is distinct from the interpolation times
$\tau_a$ and $\tau_f$ of Section~\ref{sec:training_objectives}. Each
sampled interpolation time follows
\begin{equation}
 u\sim\operatorname{Beta}(1.5,1),\qquad
 \tau=\frac{0.999-\min(u,0.999)}{0.999}.
\label{eq:noise_time}
\end{equation}
The MMDiT receives a discretized time condition obtained by scaling
$\tau$ by $1000$. Action noise is used on action-routed samples, whereas
FV auxiliary samples use the clean demonstrated action chunk and fix
$\tau_a=1$; their feature interpolation time $\tau_f$ is sampled.

\paragraph{Action sampling.}
At inference, the clean-action sampler uses four steps at
$\tau_{a,j}=j/4$, $j=0,\ldots,3$, following the interpolation convention of
Section~\ref{sec:training_objectives} and always remixing the current
estimate with the initial Gaussian noise before predicting the next clean
chunk.

\paragraph{Training budgets.}
All reported training budgets and checkpoint labels count optimizer
updates. With $g$ devices, $b$ samples per device, and gradient
accumulation factor $a$, the effective batch size in
Section~\ref{sec:predictive_targets} is $B_{\mathrm{eff}}=gba$. All
LIBERO and CALVIN runs use the 60k-update schedule in
Table~\ref{tab:launcher_defaults}; the full budget is its endpoint. The
30k, 40k, and 50k results, including the half-budget action-only baseline
in Table~\ref{tab:policy}, are intermediate checkpoints of these runs.

\paragraph{Action objectives.}
All reported policies predict clean eight-step chunks of seven-dimensional
actions, and the loss $\ell_A$ in Equation~\eqref{eq:policy_objectives}
follows the action definition of each dataset.
On LIBERO, actions are relative: the end-effector position change, the
axis-angle rotation change, and a gripper command. The continuous
dimensions are normalized with dataset statistics, the gripper remains
binary, and $\ell_A$ is the elementwise L1 error.
On CALVIN, actions are relative end-effector (TCP) commands: normalized
position changes, Euler-angle changes, and a binary gripper command
($+1$ open, $-1$ close), each mapped to $[-1,1]$ by its 1st and 99th
percentiles, and $\ell_A$ is the mean-squared error. At execution, the six
continuous outputs are clipped to $[-1,1]$, converted by the environment
to an absolute TCP target, and executed through inverse kinematics on
the seven Franka joints.
On the DOS W1 real robot, actions are six joint positions and a continuous
gripper command with the same percentile mapping, and $\ell_A$ is the
elementwise L1 error.
Within each dataset, all LA, FV, and action-only configurations therefore
share the same action loss. Validity masks exclude invalid
action timesteps, and the loss averages valid timesteps per channel before
averaging channels and samples.

\paragraph{Routing and loss weights.}
Each sample contributes only its selected loss; grouped route means are
weighted by their microbatch fractions, with the auxiliary group
additionally scaled by $\lambda_q$: $0.1$ for LAPA, $0.01$ for the two
LAPA-DINO variants, and $1$ for all FV sources. The LA weights are
selected by the ablation in Appendix~\ref{app:la_weight_ablation}; each
target-specific weight is held fixed across all allocations and
checkpoints.

\paragraph{Training data.}
One model is jointly trained on the four LIBERO suites. All LIBERO
configurations, including the action-only baseline and the source-matched
controls, use the same 1,693 training episodes, comprising 273,465 frames
over 40 tasks at 20\,Hz (Table~\ref{tab:libero_training_corpus}).
CALVIN follows the ABC$\rightarrow$D protocol: training uses environments
A, B, and C, while evaluation uses environment D. On both benchmarks, LA
augments the shared demonstrations with categorical labels, and FV targets
are computed from future observations.

\begin{table}[!htb]
\caption{LIBERO training demonstrations, shared by all configurations. The
final column gives the minimum and maximum episode counts across the ten
tasks in each suite.}
\label{tab:libero_training_corpus}
\centering
\begin{tabular*}{\linewidth}{@{\extracolsep{\fill}}lrrrr@{}}
\toprule
Suite & Tasks & Episodes & Frames & Episodes/task\\
\midrule
Spatial & 10 & 432 & 52,970 & 35--47\\
Object & 10 & 454 & 66,984 & 42--50\\
Goal & 10 & 428 & 52,042 & 33--50\\
Long & 10 & 379 & 101,469 & 29--49\\
\midrule
Total & 40 & 1,693 & 273,465 & ---\\
\bottomrule
\end{tabular*}
\end{table}

\paragraph{Optimization settings.}
We use AdamW with $\beta=(0.9,0.95)$, $\epsilon=10^{-8}$, weight decay
$10^{-8}$, and gradient-norm clipping at one.
Table~\ref{tab:launcher_defaults} lists the learning-rate schedule and the
other settings shared by the main LIBERO and CALVIN experiments; both
learning rates decay to 5\% of their peaks.

\begin{table}[!htb]
\caption{Shared optimization settings for the main LIBERO and CALVIN
experiments.}
\label{tab:launcher_defaults}
\centering
\begin{tabularx}{\linewidth}{@{}l>{\raggedright\arraybackslash}X@{}}
\toprule
Setting & Value\\
\midrule
Action horizon & 8\\
Effective batch size & 128\\
Training schedule & 60k optimizer updates; checkpoints at 30k, 40k, 50k, and 60k\\
MMDiT learning rate & Peak $10^{-4}$; final $5\times10^{-6}$\\
VLM learning rate & Peak $10^{-5}$; final $5\times10^{-7}$\\
Warmup & Linear increase from zero to peak over 2k updates\\
Decay & Cosine from update 2k to update 60k\\
\bottomrule
\end{tabularx}
\end{table}

\paragraph{Control matching.}
Within each paired LIBERO or CALVIN routing comparison, we hold fixed
model initialization, demonstration data, target encoder, action horizon,
visual inputs, auxiliary expert, action objective, effective batch size,
total optimizer updates, learning-rate schedule, and inference settings;
only the routing probability changes. At P100, every training sample
follows the action route and the objective contains only the action loss.
Each FV configuration has a source-matched P100 control that retains its
FV encoder, MMDiT input $f_t$, and two-stream architecture. The three LA
configurations instead share one action-only baseline, which keeps the
two-stream architecture and feeds the $L_z=32$ learnable LAPA queries to
the auxiliary expert. Without an LA loss, these queries are never
supervised and carry no target-specific information, so the action expert
has no LA signal to exploit, and this baseline serves as the P100
reference for all three LA targets. Cross-family comparisons may
additionally differ in target representations and auxiliary experts; the routing-only interpretation applies within a
matched configuration.
\FloatBarrier

\subsection{Policy Evaluation Protocols and Metrics}
\label{app:calvin_metrics}
\paragraph{LIBERO.}
Spatial, Object, Goal, and Long contain ten tasks each, with 50 rollouts
per task. If $y_{s,j,r}$ indicates success in rollout $r$ of task $j$
in suite $s$, then
\[
\mathrm{SR}_s=\frac{100}{10}\sum_{j=1}^{10}
\left(\frac{1}{50}\sum_{r=1}^{50}y_{s,j,r}\right),
\qquad
\mathrm{Avg.}=\frac14\sum_{s=1}^{4}\mathrm{SR}_s.
\]
Equal task and rollout counts make Avg. the pooled success rate
over 2,000 rollouts; all reported LIBERO values use evaluation seed 0.

\paragraph{CALVIN.}
All configurations use the same 1,000 ordered five-task sequences and
initial states; execution stops at the first failure. CALVIN results in
the main text use evaluation seed 0; Table~\ref{tab:calvin_three_seeds}
reports results over seeds 0, 1, and 2.
Let $L_i\in\{0,\ldots,5\}$ be the completed prefix length. Then
\begin{equation}
\mathrm{SR}\text{-}k
=\frac{100}{n}\sum_{i=1}^{n}\mathbf{1}[L_i\geq k],
\qquad k=1,\ldots,5,
\end{equation}
and the average task-chain length is
\begin{equation}
\mathrm{Avg.\ Len.}
=\frac{1}{n}\sum_{i=1}^{n}L_i
=\frac{1}{100}\sum_{k=1}^{5}(\mathrm{SR}\text{-}k).
\label{eq:calvin_metrics}
\end{equation}
SR-$k$ is cumulative rather than conditional, and Avg. Len. summarizes
the same sequence outcomes.

\begin{table}[H]
\centering\small\setlength{\tabcolsep}{3pt}\renewcommand{\arraystretch}{0.95}
\caption{CALVIN ABC$\rightarrow$D results over evaluation seeds 0, 1, and 2
(mean $\pm$ sample standard deviation); SR-5 is in \%. Half and Full denote
the half- and full-budget action-only baselines. Main-text tables and figures report seed 0.}
\label{tab:calvin_three_seeds}
\begin{minipage}[t]{0.49\linewidth}
\centering
\begin{tabular}{@{}llcc@{}}
\toprule
Target & Route & Avg. Len. & SR-5 \\
\midrule
\multirow{2}{*}{Action-only} & Half & $4.000\pm0.065$ & $58.8\pm1.7$ \\
 & Full & $4.335\pm0.040$ & $71.6\pm1.8$ \\
\midrule
\multirow{3}{*}{LAPA} & P25 & $3.501\pm0.021$ & $45.2\pm0.9$ \\
 & P50 & $3.486\pm0.059$ & $43.5\pm2.0$ \\
 & P75 & $4.092\pm0.032$ & $63.0\pm1.0$ \\
\midrule
\multirow{3}{*}{LAPA-DINOv2} & P25 & $4.084\pm0.039$ & $66.6\pm2.0$ \\
 & P50 & $3.916\pm0.051$ & $63.1\pm2.0$ \\
 & P75 & $4.308\pm0.052$ & $73.7\pm1.6$ \\
\midrule
\multirow{3}{*}{LAPA-DINOv3} & P25 & $4.267\pm0.016$ & $71.9\pm0.8$ \\
 & P50 & $4.471\pm0.034$ & $77.8\pm1.1$ \\
 & P75 & $4.395\pm0.026$ & $75.6\pm1.4$ \\
\bottomrule
\end{tabular}
\end{minipage}\hfill
\begin{minipage}[t]{0.49\linewidth}
\centering
\begin{tabular}{@{}llcc@{}}
\toprule
Target & Route & Avg. Len. & SR-5 \\
\midrule
\multirow{4}{*}{FV-DINOv3} & P25 & $2.146\pm0.040$ & $18.1\pm1.6$ \\
 & P50 & $4.365\pm0.010$ & $76.0\pm1.0$ \\
 & P75 & $4.207\pm0.053$ & $69.3\pm2.0$ \\
 & P100 & $4.245\pm0.020$ & $70.1\pm0.7$ \\
\midrule
\multirow{4}{*}{FV-V-JEPA~2} & P25 & $4.522\pm0.026$ & $80.5\pm0.6$ \\
 & P50 & $4.419\pm0.029$ & $75.0\pm1.0$ \\
 & P75 & $4.239\pm0.024$ & $72.5\pm1.0$ \\
 & P100 & $4.076\pm0.031$ & $62.8\pm0.9$ \\
\midrule
\multirow{4}{*}{FV-Wan2.2} & P25 & $4.248\pm0.016$ & $70.6\pm0.4$ \\
 & P50 & $4.254\pm0.017$ & $70.7\pm0.3$ \\
 & P75 & $4.373\pm0.010$ & $77.0\pm0.8$ \\
 & P100 & $3.753\pm0.033$ & $58.9\pm0.6$ \\
\bottomrule
\end{tabular}
\end{minipage}
\end{table}

\section{Supplementary Policy Results}
\label{app:policy}

\subsection{Primary Benchmark Details}
\label{app:primary_metric_details}
Table~\ref{tab:primary_metric_details} retains the CALVIN prefix-success
metrics omitted from the main comparison table; SR-$k$ and Avg. Len. are
defined in Equation~\eqref{eq:calvin_metrics}.
In Table~\ref{tab:policy}, Param. reports total model size rounded to the
nearest billion. Comparisons in that table do not establish a universal
ranking across training protocols.

\begin{table*}[t]
\centering
\caption{CALVIN prefix-success metrics underlying Table~\ref{tab:policy}.
SR-1 through
SR-5 are cumulative success rates (\%); Avg. Len. is measured in tasks
per sequence.}
\label{tab:primary_metric_details}
\begingroup
\normalsize
\setlength{\tabcolsep}{3pt}
\renewcommand{\arraystretch}{1.05}
\begin{tabular*}{\textwidth}{@{\extracolsep{\fill}}lrrrrrr@{}}
\toprule
Configuration & SR-1 & SR-2 & SR-3 & SR-4 & SR-5 & Avg. Len. \\
\midrule
Action-only (half budget) & 98.1 & 91.5 & 80.3 & 69.5 & 57.7 & 3.971 \\
Action-only (full budget) & 99.4 & 95.3 & 87.3 & 79.1 & 70.3 & 4.314 \\
\midrule
LAPA & 96.2 & 85.0 & 68.6 & 54.7 & 42.1 & 3.466 \\
LAPA-DINOv2 & 94.0 & 86.7 & 78.6 & 70.8 & 63.4 & 3.935 \\
LAPA-DINOv3 & 98.2 & 94.7 & 89.2 & 84.3 & 77.0 & 4.434 \\
DINOv3 FV & 97.4 & 93.8 & 88.5 & 82.2 & 75.1 & 4.370 \\
V-JEPA~2 FV & 99.0 & 96.0 & 89.9 & 84.2 & 76.0 & 4.451 \\
Wan2.2 FV & 97.0 & 92.0 & 85.2 & 79.0 & 70.5 & 4.237 \\
\bottomrule
\end{tabular*}
\endgroup
\end{table*}

\subsection{Source-Matched FV Controls}
\label{app:policy_checkpoints}
Table~\ref{tab:libero_checkpoints} compares each LIBERO P50 FV configuration
at every checkpoint with both the shared action-only baseline and its
source-matched control, which differs only in routing
(Appendix~\ref{app:protocol_records}).

\begin{table}[!htbp]
\caption{LIBERO FV comparisons with the shared action-only baseline and
source-matched controls across checkpoints. Scores are success rates (\%);
differences are P50 FV minus the indicated reference in percentage points.
All columns use the same checkpoint budget. Source-matched P100 controls
hold all settings except routing fixed.}
\label{tab:libero_checkpoints}
\centering
\setlength{\tabcolsep}{4pt}
\begin{tabular*}{\linewidth}{@{\extracolsep{\fill}}lrrrrrr@{}}
\toprule
FV source & \shortstack{Optimizer\\updates (k)} & \shortstack{Shared\\action-only} &
\shortstack{P50\\FV} & \shortstack{Source-matched\\P100} &
\shortstack{$\Delta$ shared} & \shortstack{$\Delta$ source-matched}\\
\midrule
DINOv3 & 30 & 93.20 & 91.95 & 96.65 & -1.25 & -4.70\\
 & 40 & 95.45 & 96.80 & 96.95 & +1.35 & -0.15\\
 & 50 & 96.35 & 97.55 & 97.90 & +1.20 & -0.35\\
 & 60 & 96.50 & 97.25 & 97.45 & +0.75 & -0.20\\
\midrule
V-JEPA~2 & 30 & 93.20 & 96.55 & 96.40 & +3.35 & +0.15\\
 & 40 & 95.45 & 97.10 & 97.55 & +1.65 & -0.45\\
 & 50 & 96.35 & 96.50 & 97.20 & +0.15 & -0.70\\
 & 60 & 96.50 & 97.95 & 96.85 & +1.45 & +1.10\\
\midrule
Wan2.2 & 30 & 93.20 & 97.35 & 97.15 & +4.15 & +0.20\\
 & 40 & 95.45 & 97.55 & 96.75 & +2.10 & +0.80\\
 & 50 & 96.35 & 97.10 & 97.20 & +0.75 & -0.10\\
 & 60 & 96.50 & 98.15 & 97.55 & +1.65 & +0.60\\
\bottomrule
\end{tabular*}
\end{table}

\subsection{LIBERO P50 Checkpoint Comparisons}
\label{app:libero_route05_records}
Table~\ref{tab:libero_route05_detailed} reports P50 performance at
30k, 40k, 50k, and 60k optimizer updates. These checkpoints track performance
through training under the same four-stage evaluation schedule used for the
shared action-only baseline in Table~\ref{tab:libero_checkpoints}.
Figure~\ref{fig:checkpoints}(b) in Section~\ref{sec:ablations} plots
these LIBERO trajectories, complementing the CALVIN trajectories in
Figure~\ref{fig:checkpoints}(a).

\begingroup
\setlength{\tabcolsep}{5pt}
\renewcommand{\arraystretch}{1.00}
\begin{longtable}{@{}lrrrrrr@{}}
\caption{Detailed LIBERO P50 results at four optimizer-update checkpoints.
P50 routes half of the training samples to action prediction in expectation.
Success rates are percentages; Avg. is the mean of the four suite rates.}\label{tab:libero_route05_detailed}\\
\toprule
Configuration & \shortstack{Optimizer\\updates (k)} & Spatial & Object & Goal & Long & Avg. \\
\midrule
\endfirsthead
\toprule
Configuration & \shortstack{Optimizer\\updates (k)} & Spatial & Object & Goal & Long & Avg. \\
\midrule
\endhead
\midrule
\multicolumn{7}{r}{Continued on next page}\\
\endfoot
\bottomrule
\endlastfoot
LAPA & 30 & 97.20 & 98.20 & 90.20 & 73.20 & 89.70 \\
LAPA & 40 & 97.20 & 98.80 & 94.20 & 84.00 & 93.55 \\
LAPA & 50 & 97.00 & 98.60 & 93.40 & 91.40 & 95.10 \\
LAPA & 60 & 96.40 & 98.20 & 94.20 & 89.80 & 94.65 \\
\addlinespace[3pt]
LAPA-DINOv2 & 30 & 95.80 & 97.60 & 93.60 & 84.60 & 92.90 \\
LAPA-DINOv2 & 40 & 93.60 & 97.80 & 97.40 & 87.60 & 94.10 \\
LAPA-DINOv2 & 50 & 97.00 & 98.00 & 95.80 & 92.40 & 95.80 \\
LAPA-DINOv2 & 60 & 97.00 & 99.60 & 98.00 & 91.20 & 96.45 \\
\addlinespace[3pt]
LAPA-DINOv3 & 30 & 96.60 & 97.80 & 89.20 & 77.60 & 90.30 \\
LAPA-DINOv3 & 40 & 98.40 & 99.60 & 98.00 & 90.80 & 96.70 \\
LAPA-DINOv3 & 50 & 96.60 & 97.40 & 96.40 & 90.20 & 95.15 \\
LAPA-DINOv3 & 60 & 97.20 & 97.20 & 96.80 & 92.40 & 95.90 \\
\addlinespace[3pt]
DINOv3 FV & 30 & 94.80 & 97.00 & 94.80 & 81.20 & 91.95 \\
DINOv3 FV & 40 & 98.00 & 99.80 & 97.80 & 91.60 & 96.80 \\
DINOv3 FV & 50 & 98.80 & 99.20 & 98.80 & 93.40 & 97.55 \\
DINOv3 FV & 60 & 98.40 & 99.20 & 98.40 & 93.00 & 97.25 \\
\addlinespace[3pt]
V-JEPA~2 FV & 30 & 99.00 & 98.60 & 97.00 & 91.60 & 96.55 \\
V-JEPA~2 FV & 40 & 99.00 & 99.20 & 96.80 & 93.40 & 97.10 \\
V-JEPA~2 FV & 50 & 98.40 & 99.40 & 97.40 & 90.80 & 96.50 \\
V-JEPA~2 FV & 60 & 98.80 & 99.20 & 98.60 & 95.20 & 97.95 \\
\addlinespace[3pt]
Wan2.2 FV & 30 & 99.00 & 99.80 & 97.80 & 92.80 & 97.35 \\
Wan2.2 FV & 40 & 97.80 & 99.60 & 96.60 & 96.20 & 97.55 \\
Wan2.2 FV & 50 & 98.60 & 99.20 & 96.60 & 94.00 & 97.10 \\
Wan2.2 FV & 60 & 98.40 & 100.00 & 99.00 & 95.20 & 98.15 \\
\end{longtable}
\endgroup

\subsection{LA Loss-Weight Ablation}
\label{app:la_weight_ablation}
Table~\ref{tab:la_weight_ablation} varies the LA loss weight
$\lambda_{\mathrm{LA}}\in\{0.01,0.1,1.0\}$ for each LA target on LIBERO at
P50 and the full budget, with all other settings fixed. For every target,
$\lambda_{\mathrm{LA}}=1.0$ performs worst. The two LAPA-DINO variants
perform best at $0.01$, whereas LAPA is nearly unchanged between $0.01$ and
$0.1$ and is slightly better at $0.1$. We therefore use $0.1$ for LAPA and
$0.01$ for both LAPA-DINO variants in all reported experiments; the same
weights are reused on CALVIN without further tuning.

\begin{table}[!ht]
\centering
\caption{LA loss-weight ablation on LIBERO (P50, full budget). Entries are
average success rates (Avg., \%). Bold: the
weight used for each target in all reported experiments.}
\label{tab:la_weight_ablation}
\begingroup
\small
\setlength{\tabcolsep}{8pt}
\begin{tabular}{@{}lrrr@{}}
\toprule
$\lambda_{\mathrm{LA}}$ & LAPA & LAPA-DINOv2 & LAPA-DINOv3 \\
\midrule
0.01 & 94.55 & \textbf{96.45} & \textbf{95.90} \\
0.1  & \textbf{94.65} & 94.95 & 95.50 \\
1.0  & 91.95 & 93.95 & 94.90 \\
\bottomrule
\end{tabular}
\endgroup
\end{table}

\subsection{CALVIN Checkpoint Comparisons}
\label{app:calvin_checkpoints}
Table~\ref{tab:calvin_p50_absolute} reports the absolute P50 scores underlying
Figure~\ref{fig:checkpoints}(a), enabling direct comparisons between LA and FV
at each training budget.

\begin{table}[!htbp]
\centering
\caption{CALVIN average task-chain length (Avg. Len.) under P50 routing at each tested
training budget. These absolute scores underlie the P50 trajectories in
Figure~\ref{fig:checkpoints}(a).}
\label{tab:calvin_p50_absolute}
\begin{tabular*}{\linewidth}{@{\extracolsep{\fill}}lrrrr@{}}
\toprule
Target & \multicolumn{4}{c}{Optimizer updates (k)} \\
\cmidrule(l){2-5}
 & 30 & 40 & 50 & 60 \\
\midrule
\multicolumn{5}{l}{\textit{Latent action (LA)}} \\
LAPA & 3.019 & 3.313 & 3.580 & 3.466 \\
LAPA-DINOv2 & 3.120 & 4.194 & 3.812 & 3.935 \\
LAPA-DINOv3 & 3.830 & 4.157 & 4.322 & 4.434 \\
\midrule
\multicolumn{5}{l}{\textit{Future vision (FV)}} \\
DINOv3 & 4.085 & 4.014 & 4.333 & 4.370 \\
V-JEPA~2 & 4.087 & 4.430 & 4.268 & 4.451 \\
Wan2.2 & 3.876 & 3.997 & 4.094 & 4.237 \\
\bottomrule
\end{tabular*}
\end{table}

Table~\ref{tab:calvin_checkpoints} gives P50-minus-P100 differences for
each FV source and checkpoint, using the same source-matched controls as
Figure~\ref{fig:checkpoints}(a).

\begin{table}[!htbp]
\caption{CALVIN P50 minus source-matched P100 at all four reported checkpoints.
$\Delta$ Avg. Len. is in tasks per sequence; $\Delta$ SR-5 is in percentage points.
Each comparison is positive on both metrics.}
\label{tab:calvin_checkpoints}
\centering
\begin{tabular*}{\linewidth}{@{\extracolsep{\fill}}lrrr@{}}
\toprule
FV source & \shortstack{Optimizer\\updates (k)} & $\Delta$ Avg. Len. & $\Delta$ SR-5 (pp)\\
\midrule
DINOv3 & 30 & +0.243 & +12.9\\
 & 40 & +0.235 & +9.7\\
 & 50 & +0.270 & +11.9\\
 & 60 & +0.146 & +5.0\\
\midrule
V-JEPA~2 & 30 & +0.770 & +18.2\\
 & 40 & +0.622 & +19.0\\
 & 50 & +0.258 & +10.7\\
 & 60 & +0.360 & +12.8\\
\midrule
Wan2.2 & 30 & +0.496 & +12.4\\
 & 40 & +0.702 & +17.4\\
 & 50 & +0.589 & +16.2\\
 & 60 & +0.493 & +12.2\\
\bottomrule
\end{tabular*}
\end{table}

\subsection{CALVIN Allocation Comparisons}
\label{app:calvin_allocations}
Table~\ref{tab:calvin_allocation_120k} reports every tested allocation at
60k updates: P25--P75 for each LA target and P25--P100 for each FV source.
LA uses the shared action-only baseline as P100, reported in
Table~\ref{tab:primary_metric_details}; FV uses source-matched P100 controls.

\begin{table*}[t]
\centering
\caption{Tested supervision allocations at the common
CALVIN endpoint of 60k optimizer updates. Each cell reports Avg. Len. / SR-5 (\%).
P25, P50, P75, and P100 denote action-routing probabilities
$p_A=0.25$, $0.50$, $0.75$, and $1.00$, respectively; lower P
assigns a larger expected fraction of samples to auxiliary (LA/FV) prediction.
FV P100 entries are source-matched controls; LA shares the full-budget action-only baseline in Table~\ref{tab:primary_metric_details}, hence the
dashes in its P100 column. Bold marks the best tested value for each metric
within each row, including ties.}
\label{tab:calvin_allocation_120k}
\begingroup
\normalsize
\setlength{\tabcolsep}{2.5pt}
\renewcommand{\arraystretch}{1.08}
\begin{tabular*}{\textwidth}{@{\extracolsep{\fill}}lcccc@{}}
\toprule
Target & P25 & P50 & P75 & P100 \\
\midrule
\multicolumn{5}{l}{\textit{Latent action (LA)}} \\
LAPA & 3.511 / 45.7 & 3.466 / 42.1 & \textbf{4.128 / 63.9} & --- \\
LAPA-DINOv2 & 4.126 / 68.8 & 3.935 / 63.4 & \textbf{4.264 / 72.8} & --- \\
LAPA-DINOv3 & 4.283 / 72.8 & \textbf{4.434 / 77.0} & 4.421 / \textbf{77.0} & --- \\
\addlinespace[2pt]
\multicolumn{5}{l}{\textit{Future vision (FV)}} \\
DINOv3 & 2.107 / 16.6 & \textbf{4.370 / 75.1} & 4.193 / 68.7 & 4.224 / 70.1 \\
V-JEPA~2 & \textbf{4.522 / 80.0} & 4.451 / 76.0 & 4.240 / 72.1 & 4.091 / 63.2 \\
Wan2.2 & 4.264 / 71.1 & 4.237 / 70.5 & \textbf{4.377 / 77.4} & 3.744 / 58.3 \\
\bottomrule
\end{tabular*}
\endgroup
\end{table*}

Table~\ref{tab:calvin_allocation_80k} gives the corresponding FV grid at
40k updates; Figure~\ref{fig:allocation_budget_full} covers all targets,
allocations, and budgets.

\begin{table}[!htbp]
\caption{FV supervision allocation at a fixed CALVIN training
budget of 40k optimizer updates. P25, P50, P75, and P100 denote action-routing probabilities
$0.25$, $0.50$, $0.75$, and $1.00$, respectively. P100 is the source-matched control. Batch size and optimization settings are fixed within
each source. Bold marks the best tested value for each metric within each row.}
\label{tab:calvin_allocation_80k}
\centering
\setlength{\tabcolsep}{3pt}
\begin{tabular*}{\linewidth}{@{\extracolsep{\fill}}lrrrrrrrr@{}}
\toprule
& \multicolumn{2}{c}{P25} & \multicolumn{2}{c}{P50} & \multicolumn{2}{c}{P75} & \multicolumn{2}{c}{P100}\\
\cmidrule(lr){2-3}\cmidrule(lr){4-5}\cmidrule(lr){6-7}\cmidrule(lr){8-9}
FV source & \shortstack{Avg.\\Len.} & \shortstack{SR-5\\(\%)} & \shortstack{Avg.\\Len.} & \shortstack{SR-5\\(\%)} & \shortstack{Avg.\\Len.} & \shortstack{SR-5\\(\%)} & \shortstack{Avg.\\Len.} & \shortstack{SR-5\\(\%)}\\
\midrule
DINOv3 & 1.666 & 10.3 & 4.014 & 64.1 & \textbf{4.223} & \textbf{69.7} & 3.779 & 54.4\\
V-JEPA~2 & 4.318 & 73.7 & \textbf{4.430} & \textbf{75.0} & 4.384 & 73.7 & 3.808 & 56.0\\
Wan2.2 & 4.112 & 64.4 & 3.997 & 64.3 & \textbf{4.443} & \textbf{80.0} & 3.295 & 46.9\\
\bottomrule
\end{tabular*}
\end{table}

\subsection{Real-Robot Evaluation Details}
\label{app:beyond_standard_evaluations}
Table~\ref{tab:real_robot} contains the real-robot values summarized in
Figure~\ref{fig:supporting_evaluations}.

The real-robot platform is the DOS W1 left follower arm, equipped with
third-person and wrist-mounted RealSense D435i cameras. The RGB input
resolutions are $448\!\times\!252$ and $448\!\times\!336$ pixels,
respectively, both at 15\,Hz. Actions comprise six absolute joint positions
and one continuous gripper command (action objectives in
Appendix~\ref{app:protocol_records}); depth and effort are discarded
during preprocessing.
The shared four-task training set contains 203 demonstrations: 50 each for
Pick-out, Place-in, and Rearrange-objects, and 53 for Rearrange-cups.
The four task instructions are:
\begin{enumerate}
    \item \emph{Pick the objects out of the basket.}
    \item \emph{Place the objects into the basket.}
    \item \emph{Rearrange the cups.}
    \item \emph{Rearrange the objects.}
\end{enumerate}
Auxiliary configurations use P50. All configurations are trained for 30k optimizer updates and evaluated over
20 trials per task. Each trial has a maximum score of eight. Pick-out,
Place-in, and Rearrange-objects contain four objects. For each object, a
successful pick contributes one point and a successful placement according to
the task instruction contributes one point. Rearrange-cups contains three cups:
the same pick-and-place rule contributes up to six points, and the remaining
two points evaluate whether the final arrangement is orderly and whether the
cup handles face outward from the shelf. These components sum to the trial
score, so the maximum accumulated score over 20 trials is 160 for every task.
The component-level rule allows partial credit. No scene damage occurred in
the reported trials. Results are reported separately for each task.

\begin{table}[!htbp]
\centering
\caption{Real-robot manipulation outcomes on the DOS W1 left arm.
Auxiliary configurations use P50; all configurations are trained for 30k
optimizer updates. Scores accumulate over 20 trials per task (maximum 160;
scoring in Appendix~\ref{app:beyond_standard_evaluations}). Bold marks the
highest score in each task.}
\label{tab:real_robot}
\begingroup
\normalsize
\setlength{\tabcolsep}{4pt}
\renewcommand{\arraystretch}{1.08}
\begin{tabular*}{\textwidth}{@{\extracolsep{\fill}}lcccc@{}}
\toprule
Model & \shortstack{Pick-out\\($n=20$)} & \shortstack{Place-in\\($n=20$)}
& \shortstack{Rearrange-cups\\($n=20$)}
& \shortstack{Rearrange-objects\\($n=20$)} \\
\midrule
\multicolumn{5}{l}{\textit{Action-only}} \\
Action-only baseline & 116 / 160 & 92 / 160 & 80 / 160 & 124 / 160 \\
\addlinespace[2pt]
\multicolumn{5}{l}{\textit{Latent action (LA)}} \\
LAPA (P50) & 92 / 160 & 92 / 160 & 80 / 160 & 96 / 160 \\
LAPA-DINOv2 (P50) & 112 / 160 & 80 / 160 & 76 / 160 & 120 / 160 \\
LAPA-DINOv3 (P50) & 106 / 160 & 88 / 160 & 72 / 160 & 92 / 160 \\
\addlinespace[2pt]
\multicolumn{5}{l}{\textit{Future vision (FV)}} \\
DINOv3 (P50) & 138 / 160 & 100 / 160 & 96 / 160 & 132 / 160 \\
V-JEPA~2 (P50) & 150 / 160 & 108 / 160 & 100 / 160 & 144 / 160 \\
Wan2.2 (P50) & \textbf{156 / 160} & \textbf{112 / 160}
& \textbf{108 / 160} & \textbf{148 / 160} \\
\bottomrule
\end{tabular*}
\endgroup
\end{table}

\FloatBarrier

\section{Diagnostic Protocols and Supplementary Results}
\label{app:diagnostics}
This section specifies the diagnostic targets and metrics, then reports the
measurements omitted from the main text.

\subsection{Diagnostic Protocols and Readout Targets}
\label{app:diagnostic_protocols}

\paragraph{Regression readouts and data.}
The regression splits for both datasets are taken from LARYBench
\citep{rw_lary,larybench_repo}, which re-segments AgiBotWorld-Beta
\citep{agibot_world} and CALVIN \citep{benchmark_calvin} into image-pair
regression samples. The six frozen source representations come from the
target encoders and checkpoints in Table~\ref{tab:implementation_records}.
Each source representation--readout target pair receives a separate MLP-ResNet (width
4096, two residual blocks, 50 epochs, AdamW). LAPA-style regressions use
flattened post-quantization vectors. The main ABS/delta experiment is
visual-only with frame-pair stride $s=45$; AgiBotWorld-Beta ABS predicts all
$45\times16=720$ action components. For V-JEPA~2, each sample supplies only
the two endpoint frames of its segment, resized to a short side of 256,
center-cropped to $224\times224$, and ImageNet-normalized; the target encoder
returns $196$ tokens of dimension 1024.

AgiBotWorld-Beta uses three regression splits. Seen training contains
45-step segments from 169 seen task IDs and is used to fit the readout
head. Seen validation contains held-out segments from the same 169 tasks.
It selects each probe's checkpoint by validation MSE and also serves as the
analysis split for the geometry, clustering, and delta-step5 measurements
(Appendix~\ref{app:diagnostic_geometry}). Geometry and clustering involve no
training on this split; because the delta-step5 probe is also selected on
it, the delta-step5 NMSE is a descriptive within-seen-task measure used to
compare encoders rather than a held-out estimate. Seen validation is
not episode-disjoint from seen training: the two splits share 12,239
episodes, although no identical 45-step action segment appears in both.
It therefore measures generalization within seen tasks rather
than to new trajectories. The unseen split contains 19 further tasks whose
task IDs and episodes do not overlap with the seen splits. The reported
unseen errors (Table~\ref{tab:diagnostics}) always come from the checkpoint
selected on seen validation, never from the best unseen epoch.
Table~\ref{tab:agibot_local_counts} lists the split sizes; these image-pair
samples are not independent robot trajectories. CALVIN uses the
\texttt{train\_stride5} and \texttt{val\_stride5} diagnostic splits and
reports errors on the environment-D validation split; it has no
seen/unseen task partition. These
frozen-source readouts characterize information recoverable by the tested
probes.

\begin{table}[!htb]
\caption{AgiBotWorld-Beta regression splits derived from LARYBench.
Counts are image-pair samples, not trajectories.}
\label{tab:agibot_local_counts}
\centering
\begin{tabular*}{\linewidth}{@{\extracolsep{\fill}}lrr@{}}
\toprule
Split & Task IDs & Samples\\
\midrule
Seen training & 169 & 84,162\\
Seen validation & 169 & 28,054\\
Unseen & 19 & 12,303\\
\bottomrule
\end{tabular*}
\end{table}

\paragraph{ABS and delta targets.}
ABS uses stored absolute trajectories; delta targets $\Delta_c$ are
coordinate-wise action differences, so their positional components are Cartesian
displacements. Each probe is trained on its own transformed labels. The
diagnostic setting $c$ is distinct from the policy target offset $\delta$
in Section~\ref{sec:predictive_targets}. Here,
$c=1$ uses adjacent-action differences, $c=9$ uses action differences with stride nine,
and $c=45$ denotes one difference between the segment's final and
initial actions. Each delta setting is a separately trained target, not a
metric applied to one ABS predictor. Encoder comparisons are made within
each dataset--target column.

\paragraph{The separate delta-step5 task.}
For a 45-action segment $(\mathbf a_0,\ldots,\mathbf a_{44})$,
let
\[
I_5=(i_0,\ldots,i_9)=(0,5,10,15,20,25,30,35,40,44),
\qquad
\mathbf d_j=\mathbf a_{i_j}-\mathbf a_{i_{j-1}},
\quad j=1,\ldots,9.
\]
Each $\mathbf d_j\in\mathbb R^{16}$. The regression target is
\[
\mathbf y_{\mathrm{step5}}
=\operatorname{vec}(\mathbf d_1,\ldots,\mathbf d_9)
\in\mathbb R^{144}.
\]
No temporal averaging is applied to this regression target. The last
difference spans four action steps ($40\rightarrow44$), unlike the
preceding five-step differences.

\paragraph{Normalized MSE.}
Statistics are computed from transformed training targets. For
AgiBotWorld-Beta, they are estimated per robot type and action dimension;
all nine delta-step5 positions share a dimension's statistics. Over valid
scalar entries,
\[
\operatorname{NMSE}
=\frac{1}{|\mathcal V|}
\sum_{(n,j,d)\in\mathcal V}
\bigl(\widehat{\widetilde y}_{n,j,d}
      -\widetilde y_{n,j,d}\bigr)^2,
\]
where tildes denote standardized coordinates. Delta predictions are not
integrated, and comparisons remain within each dataset--target column.

\paragraph{Millimeter position error.}
After denormalization, delta outputs are reconstructed from the ground-truth
initial action at endpoints $i_1,\ldots,i_M$:
\[
\widehat{\mathbf a}_{i_j}
=\mathbf a_0+\sum_{k=1}^{j}\widehat{\mathbf d}_k.
\]
For AgiBotWorld-Beta, the millimeter Euclidean position error averages arms,
evaluated times, and samples:
\[
E_{\mathrm{mm}}
=\frac{1000}{N}\sum_{n=1}^{N}
\frac{1}{|\mathcal T_n|}\sum_{t\in\mathcal T_n}
\frac{1}{2}\sum_{e\in\{\mathrm L,\mathrm R\}}
\left\|\widehat{\mathbf p}_{n,t,e}-\mathbf p_{n,t,e}\right\|_2.
\]
For CALVIN, which has a single arm, $E_{\mathrm{mm}}$ omits the average
over arms. For delta-step5, $\mathcal T_n=\{5,10,15,20,25,30,35,40,44\}$.
The initial action is an evaluation boundary condition, not a probe input;
the score is neither coordinate-wise nor a per-increment error. ABS outputs
are scored directly, without this ground-truth initial-action condition.
Thus, ABS and delta errors evaluate different readout tasks.

Table~\ref{tab:diagnostics} in the main text reports the resulting position
errors for each dataset and readout target.

\paragraph{Geometry.}
Geometry uses mean-pooled features from 28k AgiBotWorld-Beta
seen-validation samples and averages the nine delta-step5 differences:
\[
\overline{\mathbf d}
=\frac{1}{9}\sum_{j=1}^{9}\mathbf d_j
=\frac{\mathbf a_{44}-\mathbf a_0}{9}.
\]
This 16-dimensional descriptor captures net change, not the time-resolved
regression target (which it complements rather than replicates), a unit
direction, or per-time velocity. Distances use subspaces of the 16 action
channels:
\[
\texttt{pos6}=[l_x,l_y,l_z,r_x,r_y,r_z],\qquad
\texttt{posgrip8}=[\texttt{pos6},r_g,l_g],
\]
that is, the 3D end-effector displacements of both arms, optionally
augmented with both gripper channels. $S^{\mathrm{dist}}_0$ and
$S^{\mathrm{trust}}_0$ use \texttt{posgrip8}; clustering uses
\texttt{pos6}. The eight quaternion channels are excluded: the descriptor
subtracts actions componentwise, and quaternions are sign-ambiguous and
non-Euclidean, so Euclidean distances between them would introduce
spurious differences.
For randomly sampled pairs of examples, we compute Euclidean distances
between their action descriptors and between their latent representations.
The pair-distance score $S^{\mathrm{dist}}_0$ is the Pearson correlation
between these two sets of distances. Higher correlation indicates more
consistent global distance structure across the two spaces.

The trustworthiness score $S^{\mathrm{trust}}_0$ measures whether nearest
neighbors in latent space are also nearby in action space. It penalizes
latent-space neighbors that are not action-space neighbors according to
their ranks in action space.
The linear positional readout is summarized by $R^2_{\mathrm{pos}}$.
Correlations between these scores and delta-step5 NMSE are computed across
the six encoders, separately from the sample-pair distance correlations
within each encoder.

\paragraph{Clustering.}
Clustering uses mean-pooled features on seen validation, one sample
per episode, and K-means with $K=200$ feature clusters. Action
pseudo-labels are obtained separately by clustering ground-truth delta
displacements in \texttt{pos6}. Normalized mutual information (NMI) compares feature-cluster
assignments with these displacement-derived labels and with dataset
task IDs.

\subsection{Geometry and Task--Action Clustering}
\label{app:diagnostic_geometry}
Table~\ref{tab:diagnostic_geometry} expands Figure~\ref{fig:clusters}a.
Beyond the Pearson correlations in its bottom row, trustworthiness has a
Spearman coefficient of $-0.93$ with delta-step5 NMSE. These descriptive
associations involve offline readout error, not policy performance.
\begin{table}[!htb]
\caption{Feature--action geometry and separate delta-step5 regression
on AgiBotWorld-Beta seen validation (protocol in
Appendix~\ref{app:diagnostic_protocols}). Regression reports normalized MSE
(NMSE). Bold marks the highest geometry score and lowest regression
error. The bottom-row Pearson correlations relate each geometry/readout
score to NMSE across $N=6$ encoders.}
\label{tab:diagnostic_geometry}
\centering
\begin{tabular*}{\linewidth}{@{\extracolsep{\fill}}lrrrr@{}}
\toprule
Encoder & $S^{\mathrm{dist}}_0$ & $S^{\mathrm{trust}}_0$
& $R^2_{\mathrm{pos}}$ & $\mathrm{NMSE}_{\text{delta-step5}}$\\
\midrule
LAPA-DINOv3 & \textbf{0.375} & \textbf{0.763} & \textbf{0.354} & \textbf{0.807}\\
LAPA-DINOv2 & 0.189 & 0.628 & 0.190 & 0.809\\
V-JEPA~2 & -0.030 & 0.523 & 0.240 & 0.858\\
DINOv3 & 0.047 & 0.525 & -0.027 & 0.893\\
Wan2.2 & -0.025 & 0.511 & -0.001 & 0.959\\
LAPA & -0.089 & 0.479 & 0.004 & 0.959\\
\midrule
Reported Pearson $r$ vs.\ NMSE & $-0.820$ & $-0.812$ & $-0.846$ & --\\
\bottomrule
\end{tabular*}
\end{table}

Table~\ref{tab:diagnostic_clustering} expands Figure~\ref{fig:clusters}b and
reports complementary associations with displacement-derived action labels
and task IDs.
\begin{table}[!htb]
\caption{Cluster associations on
AgiBotWorld-Beta seen validation (protocol in
Appendix~\ref{app:diagnostic_protocols}). NMI compares feature clusters
with displacement-derived action labels and with task IDs.}
\label{tab:diagnostic_clustering}
\centering
\begin{tabular*}{\linewidth}{@{\extracolsep{\fill}}lrrr@{}}
\toprule
Encoder & $\mathrm{NMI}_{\mathrm{act}}$ & $\mathrm{NMI}_{\mathrm{task}}$
& Task purity\\
\midrule
LAPA-DINOv3 & 0.412 & 0.221 & 0.068\\
LAPA-DINOv2 & 0.232 & 0.174 & 0.045\\
LAPA & 0.143 & 0.148 & 0.032\\
DINOv3 & 0.183 & 0.828 & 0.653\\
V-JEPA~2 & 0.191 & 0.811 & 0.663\\
Wan2.2 & 0.155 & 0.577 & 0.370\\
\bottomrule
\end{tabular*}
\end{table}

\FloatBarrier

\subsection{Policy Attention Maps}
\label{app:policy_attention}
Both rows of Figure~\ref{fig:policy_attention} share one extraction
protocol and use the seven full-budget policies of
Table~\ref{tab:policy}. The maps are read from the action expert's
cross-attention to the VLM conditioning sequence $h_t$, not from attention
inside the frozen target encoders. We use the last quarter of the 16 MMDiT
layers (layers 13--16) and average all 12 heads per layer and the four
layers with equal weights. Queries are the eight action tokens; the state
token is excluded. Attention is taken at the final update of the four-step
action sampler ($\tau_a=3/4$; Appendix~\ref{app:protocol_records}) and
averaged over three inference noise seeds. The softmax is computed over the
full valid conditioning sequence; we then keep the entries on visual
tokens and normalize them to a density within each camera view.

\FloatBarrier

\fi
\end{document}